\documentclass{article} 
\usepackage{iclr2027_conference,times}

\usepackage{amsmath,amsfonts, bm}

\def\eqref#1{equation~\ref{#1}}

\def\1{\bm{1}}

\DeclareMathAlphabet{\mathsfit}{\encodingdefault}{\sfdefault}{m}{sl}
\SetMathAlphabet{\mathsfit}{bold}{\encodingdefault}{\sfdefault}{bx}{n}

\usepackage{amsmath}
\usepackage{amssymb}
\usepackage{mathtools}
\usepackage{graphicx}
\usepackage{subcaption}
\usepackage{booktabs}
\usepackage{multirow}
\usepackage{tabularx}
\usepackage{array}
\usepackage[table]{xcolor}
\definecolor{oursgray}{gray}{0.94}
\usepackage{xspace}
\usepackage{xcolor}
\usepackage{hyperref}
\usepackage{url}
\usepackage[nameinlink,capitalize,noabbrev]{cleveref}

\crefname{figure}{Fig.}{Figs.}
\Crefname{figure}{Fig.}{Figs.}
\crefname{table}{Table}{Tables}
\Crefname{table}{Table}{Tables}
\crefname{section}{Sec.}{Secs.}
\Crefname{section}{Sec.}{Secs.}
\crefname{subsection}{Sec.}{Secs.}
\Crefname{subsection}{Sec.}{Secs.}
\crefname{equation}{Eq.}{Eqs.}
\Crefname{equation}{Eq.}{Eqs.}

\usepackage[most]{tcolorbox}
\usepackage{enumitem}

\definecolor{promptframe}{HTML}{4F7C78}
\definecolor{promptbg}{HTML}{F6FAF9}
\definecolor{prompttitle}{HTML}{FFFFFF}
\definecolor{promptsep}{HTML}{CFE0DE}

\newtcolorbox{PromptBoxInner}[1]{%
  enhanced,
  colback=promptbg,
  colframe=promptframe,
  coltitle=prompttitle,
  fonttitle=\bfseries\footnotesize,
  title={#1},
  boxrule=0.6pt,
  arc=2pt,
  left=7pt, right=7pt, top=4pt, bottom=4pt,
  toptitle=3pt, bottomtitle=3pt,
  fontupper=\footnotesize,
  before upper={\setlength{\parskip}{0.4em}\setlength{\parindent}{0pt}},
}

\newenvironment{PromptBox}[3]{%
  \begin{figure*}[t]
  \centering
  \def\PromptBoxCaption{#2}%
  \def\PromptBoxLabel{#3}%
  \begin{minipage}{0.96\textwidth}
  \begin{PromptBoxInner}{#1}
}{%
  \end{PromptBoxInner}
  \end{minipage}
  \caption{\PromptBoxCaption}
  \label{\PromptBoxLabel}
  \end{figure*}
}

\title{Learning the Loop, Not Just the Page: Execution-Grounded Loop Learning for Web Generation}

\author{%
  Yuxin Meng$^{1,2*}$,
  Ruixu Zhang$^{1*}$,
  Junjie Wang$^{1*}$,
  Yuhan Suo$^{1,2}$,
  Yuhan Sun$^{3}$ \\
  {\bf
  Ruining Hu$^{4}$,
  Yiyao Yu$^{1}$,
  Yubin Wang$^{2}$,
  Shouwei Ruan$^{5}$,
  Bin Wang$^{2}$,
  Yue Liao$^{6}$
  } \\
  {\bf
  Yuxiang Zhang$^{2\dagger}$,
  Yujiu Yang$^{1\dagger}$
  } \\
  \textsuperscript{1}Tsinghua University \enspace
  \textsuperscript{2}Huawei Noah's Ark Lab \enspace
  \textsuperscript{3}East China Normal University \\
  \textsuperscript{4}Tongji University \enspace
  \textsuperscript{5}Institute of Artificial Intelligence, Beihang University \\
  \textsuperscript{6}National University of Singapore \\
  {\small
  \{\href{mailto:meng-yx25@mails.tsinghua.edu.cn}{meng-yx25},%
  \href{mailto:rx-zhang25@mails.tsinghua.edu.cn}{rx-zhang25}\}@mails.tsinghua.edu.cn \quad
  \href{mailto:wangjunjie@sz.tsinghua.edu.cn}
       {wangjunjie@sz.tsinghua.edu.cn}
  } \\
  {\small
  \href{mailto:joel0495@asagi.waseda.jp}
       {joel0495@asagi.waseda.jp} \quad
  \href{mailto:yang.yujiu@sz.tsinghua.edu.cn}
       {yang.yujiu@sz.tsinghua.edu.cn}
  } \\
  \multicolumn{1}{@{}l@{}}{%
    \makebox[\textwidth][c]{\url{https://github.com/IIGROUP/WebLoop}}%
  }
}

\newcommand{\nbf}[1]{\noindent\textbf{#1}}

\makeatletter
\newcommand{\blfootnote}[1]{%
  \begingroup
  \renewcommand{\thefootnote}{}%
  \long\def\@makefntext##1{\noindent##1}%
  \footnotetext[0]{#1}%
  \endgroup
}
\makeatother

\iclrfinalcopy 
\begin{document}

\maketitle
\ificlrfinal
  \fancyhead{}
  \renewcommand{\headrulewidth}{0.4pt}

\blfootnote{%
  \makebox[1em][l]{$^{*}$}Equal contribution.\\
  \makebox[1em][l]{$^{\dagger}$}Corresponding authors.\\
  \makebox[1em][l]{}Under Review.
}
\else
  \hypersetup{pdfauthor={}}
\fi

\begin{abstract}
Functional Web generation is increasingly optimized with executable rewards, yet existing methods largely focus on the quality of the final page and leave the process of diagnosing and repairing imperfect implementations underexplored.
We identify a central challenge in this setting: the Generator and Refiner produce executable artifacts with direct environment rewards, whereas the intermediate Critic influences downstream behavior without a directly executable outcome.
We introduce \textbf{WebLoop}, an execution-grounded framework that jointly learns generation, critique, and refinement within a shared policy.
WebLoop trains an execution-free Critic with complementary signals for requirement-level discriminability and downstream helpfulness, first establishing reliable diagnosis and then introducing consequence-aware credit, while all three roles are jointly optimized with group-relative policy learning.
With Qwen3.5-9B, WebLoop reaches 41.5 Overall on WebRise and 38.9\% accuracy on WebGen-Bench, improving the base model by 11.3 and 15.4 points, respectively.
The gains transfer to first-pass generation, persist at 27B scale, and generalize from text-only training to multimodal inputs.
Controlled analyses further show that the improvement cannot be explained by an additional refinement pass alone, highlighting the importance of learning the Critic and the loop itself.
\end{abstract}

\section{Introduction}

Large language models are rapidly moving Web generation from page-level code synthesis toward \emph{functional Web generation}, where models must not only produce complete HTML/CSS/JavaScript but also satisfy natural-language requirements through correct interactions, state transitions, and persistent behaviors~\citep{lu2025webgenbench,meng2026webrise}.
Unlike conventional code or text generation, generated Web pages are directly executable, allowing functional correctness to be verified through browser interactions and further optimized with reinforcement learning~\citep{jiang2026webgenr1,chen2026webgrader}.
However, existing execution-based learning remains largely outcome-centric: it optimizes what the model ultimately generates, while paying considerably less attention to how the model should diagnose and repair an already imperfect implementation.
In practical Web development, first-pass generation is rarely the end of the process, and \emph{reliable diagnosis and revision are equally critical to producing a functional page}.

\begin{figure}[t]
\centering
\includegraphics[width=\textwidth]{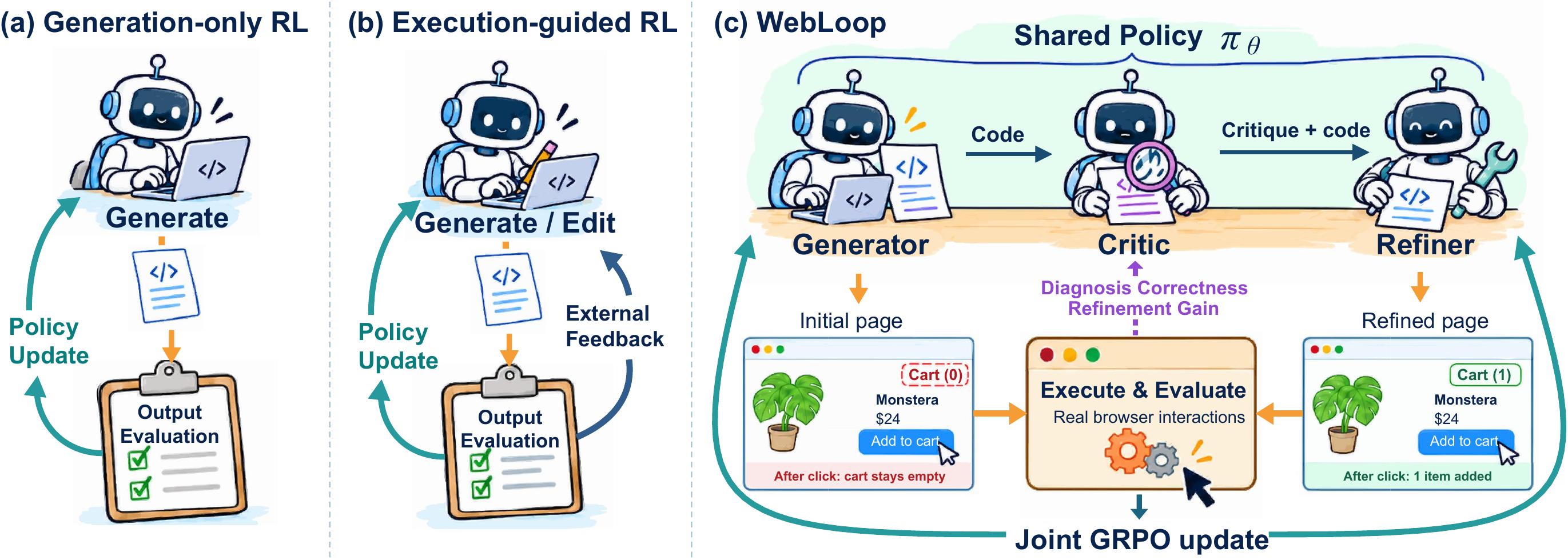}
\caption{
\textbf{Conceptual comparison of execution-based learning paradigms for functional Web generation.}
(a) Generation-only RL optimizes executable outputs directly.
(b) Execution-guided RL feeds external feedback back into the code policy.
(c) WebLoop jointly learns Generator, Critic, and Refiner, assigning execution-derived credit to the intermediate Critic.
}
\label{fig:intro}
\end{figure}

A natural direction is self-critique and refinement, where models inspect their own outputs and use the resulting feedback to guide subsequent revisions~\citep{madaan2023selfrefine,shinn2023reflexion,gou2023critic,chen2024selfdebug}, with related paradigms increasingly adopted in Web generation~\citep{lu2025webgenagent,li2026relook}.
As shown in~\cref{fig:intro}, execution supervision is typically used either to directly optimize generated Web pages, as in generation-only RL (\cref{fig:intro}(a)), or as external feedback for iterative editing (\cref{fig:intro}(b)).
Neither formulation directly addresses how to optimize an intermediate Critic that guides subsequent refinement.
From a reinforcement learning perspective, this introduces an \emph{intermediate credit-assignment problem}: the Generator and Refiner produce executable artifacts and receive direct environment rewards, whereas the Critic influences downstream behavior without a directly executable outcome.
An effective Critic must therefore learn both \emph{discriminability}, which identifies satisfied and violated requirements, and \emph{helpfulness}, which reflects whether its feedback improves subsequent refinement.
Since accurate diagnosis does not necessarily lead to effective repair, the central challenge is how to assign reliable RL credit to the Critic from verifiable execution outcomes.

To address this challenge, we introduce \textbf{WebLoop}, an execution-grounded loop learning framework shown in~\cref{fig:intro}(c).
WebLoop jointly optimizes Generator, Critic, and Refiner as interdependent roles of a shared policy.
The Generator first produces a candidate page and receives functional rewards from browser execution.
The Critic then evaluates the task requirements and candidate source code without access to test results or execution traces.
Conditioned on this critique, the Refiner revises the implementation and is evaluated in the same browser environment.
WebLoop trains the Critic with two complementary signals: requirement-level execution labels supervise its discriminability, while the execution improvement induced by subsequent refinements measures its helpfulness.
We first establish reliable diagnostic behavior and then introduce downstream improvement for consequence-aware optimization, while all three roles form separate comparison groups for group-relative advantage estimation and are jointly optimized through GRPO.
The Critic therefore remains \emph{execution-free} during inference while receiving execution-derived retrospective credit during training, turning a fixed self-refinement pipeline into a learnable RL loop.

We evaluate WebLoop on functional Web generation benchmarks across model scales and input modalities.
With Qwen3.5-9B, WebLoop reaches 41.5 Overall on WebRise and 38.9\% accuracy on WebGen-Bench, improving the corresponding base model by 11.3 and 15.4 points, respectively.
The gains transfer to first-pass generation, persist at the 27B scale, and generalize from text-only training to Markdown, Sketch, Image, and Video inputs.
Controlled ablations further show that the improvement cannot be attributed to an additional refinement pass alone: joint loop optimization consistently outperforms staged alternatives, while the Critic benefits from complementary discriminability and downstream helpfulness signals with staged credit assignment.

Our contributions are threefold.
(1) We formulate functional Web self-refinement as a structured reinforcement learning problem and identify intermediate Critic credit assignment as a central challenge under asymmetric reward observability.
(2) We propose WebLoop, which learns an execution-free Critic through complementary discriminability and helpfulness signals and jointly optimizes the Generator, Critic, and Refiner through group-relative policy learning.
(3) We conduct controlled evaluations across Web benchmarks, model scales, and input modalities, demonstrating consistent gains over generation-only and refinement-only alternatives and validating the role of Critic learning and loop-level optimization.

\section{Related Work}
\label{sec:related_work}

\nbf{Functional Web Generation and Execution-Based Learning.}
Web generation has evolved from reconstructing interfaces from screenshots or visual specifications toward generating complete and interactive Web applications.
Early and concurrent work studies large-scale screenshot-to-code learning and front-end reconstruction through datasets and benchmarks such as WebSight, Design2Code, and Web2Code~\citep{websight,design2code,web2code}, while FrontendBench further emphasizes executable evaluation for realistic front-end development~\citep{frontendbench}.
More recent benchmarks shift the focus from visual fidelity toward functional correctness, using interaction traces, state transitions, and requirement-level tests to evaluate generated websites~\citep{lu2025webgenbench,meng2026webrise}.
This verifiability has also enabled reinforcement learning for Web generation.
WebGen-R1 and WebGrader optimize Web policies with executable rewards or programmatic graders~\citep{jiang2026webgenr1,chen2026webgrader}, while WebGen-Agent incorporate iterative interaction or rendering feedback into step-wise Web optimization~\citep{lu2025webgenagent}.
These methods establish execution as an effective signal for optimizing Web artifacts.
In contrast, WebLoop further uses executable outcomes to assign learning credit to the intermediate critique that determines how an imperfect artifact is subsequently revised.

\nbf{Critique-Guided Refinement and Self-Improvement.}
A broad line of work improves model outputs through self-generated feedback, reflection, verification, or tool-assisted correction~\citep{madaan2023selfrefine,shinn2023reflexion,gou2023critic,chen2024selfdebug}.
Since intrinsic self-correction is often unreliable, later work explicitly trains correction, evaluation, or critic behavior with supervised or reinforcement signals~\citep{selfcorrect_limit,score,criticgpt,selftaught}.
More recent methods optimize critiques according to the quality of their induced revisions, with RCO~\citep{rco} and Critique-RL~\citep{Critique-RL} directly connecting critic learning to downstream refinement and staged critic optimization.
Similar ideas have entered Web generation, where WebGen-Agent~\citep{lu2025webgenagent} and ReLook~\citep{li2026relook} use external or multimodal feedback to support iterative revision.
WebLoop instead learns an execution-free Critic from requirement-level diagnosis and downstream execution gains, while jointly optimizing Generator, Critic, and Refiner within a shared policy.

\section{WebLoop: Learning the Generation--Critique--Refinement Loop}
\label{sec:method}

\begin{figure}[!t]
\centering
\includegraphics[width=1.0\textwidth]{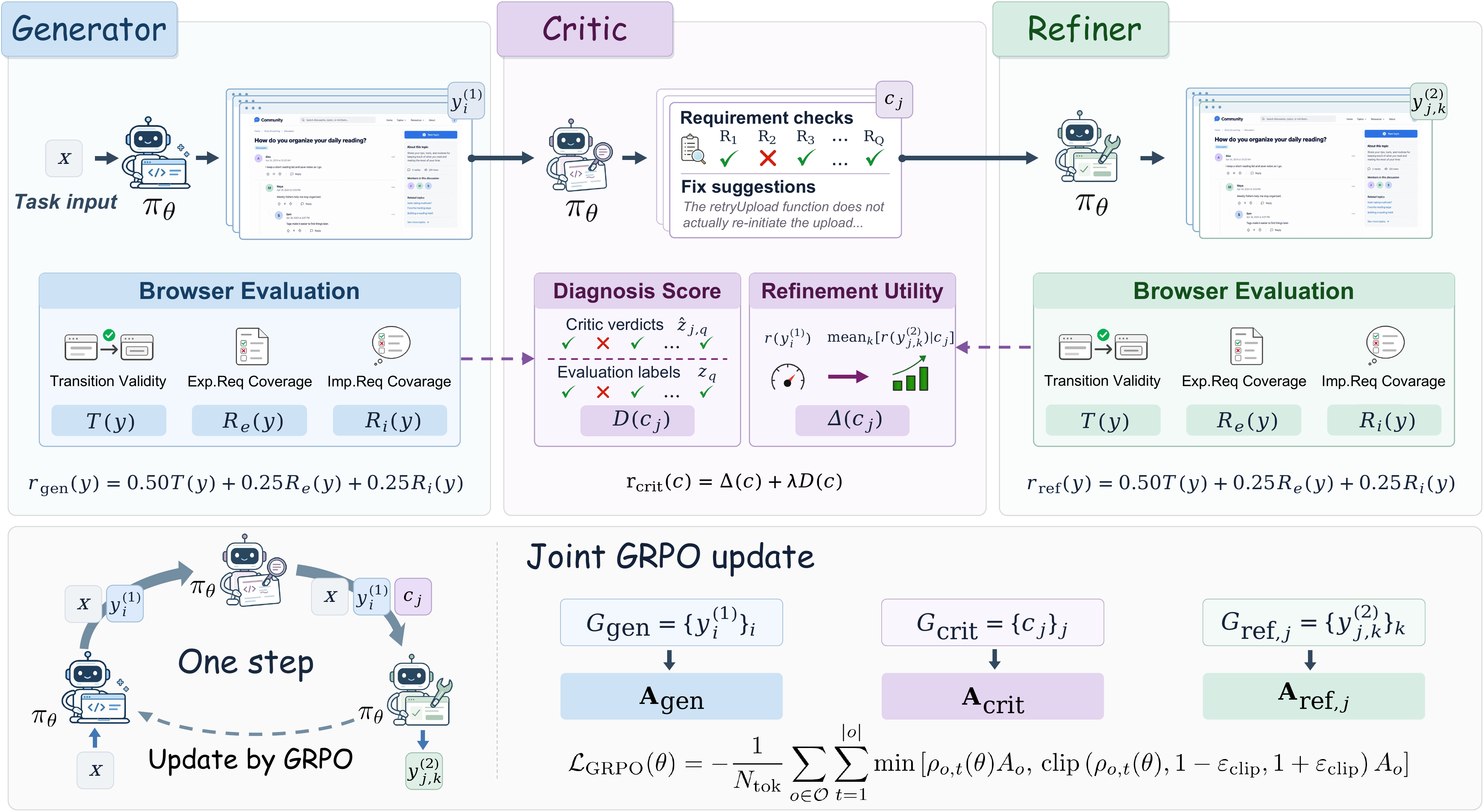}
\caption{
\textbf{Overview of our WebLoop framework.}
A shared policy sequentially acts as Generator, Critic, and Refiner.
The Generator and Refiner produce executable Web pages evaluated in the same browser environment, while the execution-free Critic guides subsequent refinement and receives training credit from its diagnosis and downstream effects.
}
\label{fig:pipeline}
\end{figure}

As shown in~\cref{fig:pipeline}, WebLoop organizes generation, critique, and refinement into a sequential loop implemented by a shared policy.
Executable outputs are directly grounded by browser evaluation, while the intermediate critique conditions subsequent refinement.

\subsection{Loop Formulation and Execution Rewards}
\label{sec:loop_formulation}

\nbf{Generator--Critic--Refiner loop.}
Let $x=(d,\mathcal R_x)$ denote a Web generation task, where $d$ is the natural-language description and $\mathcal R_x=\{r_q\}_{q=1}^{Q_x}$ contains its functional requirements.
WebLoop uses a shared policy $\pi_\theta$ to instantiate three roles under different conditioning contexts.
The Generator maps the task to a complete initial HTML implementationThe Critic inspects an imperfect implementation against the task requirements and produces requirement-level judgments and repair guidance.
The Refiner conditions on the task, the current implementation, and the critique to generate a complete revised HTML implementation.

Let $p_{\mathrm{gen}}$, $p_{\mathrm{crit}}$, and $p_{\mathrm{ref}}$ denote the corresponding prompt construction functions, and let $\pi_{\theta_{\mathrm{old}}}$ denote the rollout policy.
For each task, WebLoop samples the three stages as
\begin{equation}
\small
\begin{aligned}
y_i^{(1)}
&\sim
\pi_{\theta_{\mathrm{old}}}
\big(
\cdot \mid p_{\mathrm{gen}}(x)
\big),
&& i=1,\ldots,n_g,
\\
c_j
&\sim
\pi_{\theta_{\mathrm{old}}}
\big(
\cdot \mid p_{\mathrm{crit}}(x,y_b)
\big),
&& j=1,\ldots,n_c,
\\
y_{j,k}^{(2)}
&\sim
\pi_{\theta_{\mathrm{old}}}
\big(
\cdot \mid p_{\mathrm{ref}}(x,y_b,c_j)
\big),
&& k=1,\ldots,n_r,
\end{aligned}
\label{eq:loop-rollout}
\end{equation}
where $y_b$ denotes the imperfect base implementation selected from the initial generations.
The Generator and Refiner therefore produce executable artifacts, whereas the Critic produces an intermediate feedback decision that changes the conditional distribution of subsequent refinements.

\nbf{Execution reward.}
Both initial and refined implementations are evaluated by the same browser environment.
Following WebRISE~\citep{meng2026webrise}, each task is associated with an Interaction Contract Graph $\mathcal G_x$ that specifies browser interactions and functional assertions.
Executing a page $y$ returns
\begin{equation}
\small
\mathcal E(y;\mathcal G_x)
=
\big(
T(y),
R_{\mathrm e}(y),
R_{\mathrm i}(y),
\mathbf z(y)
\big),
\label{eq:browser-evaluation}
\end{equation}
where $T(y)$ denotes transition validity, $R_{\mathrm e}(y)$ and $R_{\mathrm i}(y)$ denote explicit and implicit requirement coverage, and $\mathbf z(y)\in\{0,1\}^{Q_x}$ contains requirement-level execution labels.
We define the page-level execution reward as
\begin{equation}
\small
r_{\mathrm{web}}(y)
=
0.50T(y)
+
0.25R_{\mathrm e}(y)
+
0.25R_{\mathrm i}(y).
\label{eq:web-reward}
\end{equation}
Both Generator and Refiner outputs are optimized with $r_{\mathrm{web}}$, placing the two executable endpoints in the same reward space.
This shared objective allows the behavioral change before and after critique to be measured directly.

\nbf{Critique-conditioned branching.}
After evaluating the initial generations, we select the median-reward candidate among valid but imperfect implementations as the shared base draft $y_b$.
All critiques are sampled from the same $(x,y_b)$ context, which removes variation caused by different starting implementations.
Importantly, the Critic observes only the task requirements and the source code of $y_b$, without access to browser results or execution traces.
Each critique $c_j$ then conditions $n_r$ independent refinements, and every refined page is evaluated under the same execution objective.
Multiple refinements under the same critique provide a more robust estimate of how effectively the current Refiner can act on that feedback.
If no valid imperfect draft exists, the task contributes only its initial-generation samples.
Further details are provided in~\cref{app:objective_details}.

This construction gives direct environment rewards to the Generator and Refiner, while the intermediate Critic has no directly executable outcome.
We next assign Critic credit using requirement-level diagnosis and downstream refinement effects.

\subsection{Credit Assignment for the Critic}
\label{sec:critic_credit}

Unlike the executable Generator and Refiner outputs, a critique has no direct browser outcome from which to obtain an environment reward.
Following the critic decomposition in Critique-RL~\citep{Critique-RL}, we distinguish two complementary properties of a critique: \emph{discriminability}, which measures whether it correctly diagnoses the current implementation, and \emph{helpfulness}, which measures whether acting on it improves downstream execution.
WebLoop grounds these two properties with requirement-level browser labels and executable refinement outcomes, respectively.

\nbf{Execution-free critique.}
For the selected base draft $y_b$, the Critic receives only the task requirements and source code through $p_{\mathrm{crit}}(x,y_b)$.
It predicts whether each explicit requirement is satisfied and provides corresponding repair guidance.
Importantly, browser results, requirement labels, and execution traces are never exposed to the Critic as input.
Execution is instead used only to construct training signals, requiring the Critic to infer functional failures directly from the specification and implementation.

\nbf{Diagnostic discriminability.}
Let $z_q\in\{0,1\}$ denote the execution-grounded label for requirement $r_q$, where $z_q=1$ only if all corresponding assertions are satisfied.
For critique $c_j$, let $\hat z_{j,q}$ denote its predicted requirement status.
We define the failed and satisfied requirement sets as
\begin{equation}
\small
\mathcal F=\{q:z_q=0\},
\qquad
\mathcal P=\{q:z_q=1\}.
\label{eq:requirement-classes}
\end{equation}
To avoid bias toward the majority class, we measure diagnostic discriminability using balanced class-wise accuracy,
\begin{equation}
\small
D(c_j)
=
a_{\mathrm{fail}}(c_j)
+
a_{\mathrm{pass}}(c_j)
-
1,
\label{eq:diagnosis-score}
\end{equation}
where $a_{\mathrm{fail}}$ and $a_{\mathrm{pass}}$ denote the prediction accuracies over $\mathcal F$ and $\mathcal P$, respectively.
When only one class is present, we linearly rescale its accuracy to the same $[-1,1]$ range.
Thus, $D(c_j)$ measures whether the Critic correctly distinguishes satisfied from violated requirements independently of downstream refinement.

\nbf{Retrospective helpfulness.}
Diagnostic correctness alone does not indicate whether a critique can actually guide an effective repair.
For each critique $c_j$, we therefore estimate its downstream utility from the $n_r$ refinements conditioned on it:
\begin{equation}
\small
H(c_j)
=
\frac{1}{n_r}
\sum_{k=1}^{n_r}
r_{\mathrm{web}}\big(y_{j,k}^{(2)}\big)
-
r_{\mathrm{web}}(y_b).
\label{eq:critique-helpfulness}
\end{equation}
A positive $H(c_j)$ indicates that refinements conditioned on the critique outperform the original implementation on average.
Using multiple refinements reduces sensitivity to a single stochastic regeneration and measures the practical utility of the critique under the current Refiner.
While $D(c_j)$ captures what the Critic understands, $H(c_j)$ captures what happens when its feedback is acted upon.

\nbf{Staged critic credit.}
Inspired by the two-stage optimization principle of Critique-RL~\citep{Critique-RL}, we first establish reliable discriminability before introducing downstream helpfulness.
This separation is particularly important in WebLoop because helpfulness is a delayed signal that depends jointly on the critique and the current Refiner, whereas discriminability can be directly grounded by requirement-level execution labels.
At update step $m$, we define the Critic reward as
\begin{equation}
r_{\mathrm{crit}}^{(m)}(c_j)
=
\begin{cases}
D(c_j), & m\leq 15,\\
H(c_j)+\lambda D(c_j), & m>15,
\end{cases}
\label{eq:staged-critic-reward}
\end{equation}
where $\lambda$ controls the contribution of diagnostic discriminability in the second stage.
Retaining $D(c_j)$ prevents downstream improvement from serving as a substitute for correct diagnosis, while $H(c_j)$ encourages feedback that can be effectively acted upon by the Refiner.
This schedule prevents downstream improvement from serving as a substitute for correct diagnosis while gradually introducing consequence-aware credit.
The resulting scalar reward allows critique samples to participate in policy optimization while preserving their distinct reward semantics from executable Web outputs.
Complete reward definitions are provided in~\cref{app:objective_details}.

\subsection{Joint Group-Relative Optimization}
\label{sec:joint_optimization}

WebLoop produces three types of policy outputs with different conditioning contexts and reward semantics.
We therefore construct role specific comparison groups for advantage estimation, and then jointly optimize all outputs under the shared policy.

\nbf{Role specific comparison groups.}
For each task, Generator outputs are compared under the same task, Critic outputs under the same base draft, and Refiner outputs under the same critique.
Specifically, we define
\begin{equation}
\small
G_{\mathrm{gen}}(x)
=
\{y_i^{(1)}\}_{i=1}^{n_g},\quad
G_{\mathrm{crit}}(x,y_b)
=
\{c_j\}_{j=1}^{n_c},\quad
G_{\mathrm{ref},j}(x,y_b,c_j)
=
\{y_{j,k}^{(2)}\}_{k=1}^{n_r}.
\label{eq:role-specific-groups}
\end{equation}
The corresponding rewards are $r_{\mathrm{web}}$ for Generator and Refiner outputs and $r_{\mathrm{crit}}$ for Critic outputs.
For any output $o$ with comparison group $g(o)$, we compute the group relative advantage as
\begin{equation}
\small
A_o
=
\frac{
r(o)-\mu_{g(o)}
}{
\sigma_{g(o)}+\epsilon_{\mathrm{norm}}
},
\label{eq:group-relative-advantage}
\end{equation}
where $\mu_{g(o)}$ and $\sigma_{g(o)}$ are the reward mean and standard deviation within its group.
This grouping keeps each comparison local to outputs generated under the same decision context, while preserving the distinct credit semantics of the three roles.

\nbf{Joint policy update.}
After advantage estimation, Generator, Critic, and Refiner samples are merged into a single optimization batch $\mathcal O$.
For an output $o$ with conditioning context $u_o$, the token level probability ratio is
\begin{equation}
\small
\rho_{o,t}(\theta)
=
\frac{
\pi_\theta(o_t\mid u_o,o_{<t})
}{
\pi_{\theta_{\mathrm{old}}}(o_t\mid u_o,o_{<t})
}.
\label{eq:token-probability-ratio}
\end{equation}
We optimize the shared policy with the clipped GRPO objective
\begin{equation}
\small
\mathcal L_{\mathrm{GRPO}}(\theta)
=
-\frac{1}{N_{\mathrm{tok}}}
\sum_{o\in\mathcal O}
\sum_{t=1}^{|o|}
\min\Big(
\rho_{o,t}(\theta)A_o,
\operatorname{clip}
\big(
\rho_{o,t}(\theta),
1-\epsilon_{\mathrm{clip}},
1+\epsilon_{\mathrm{clip}}
\big)A_o
\Big),
\label{eq:joint-grpo-loss}
\end{equation}
where $N_{\mathrm{tok}}$ is the number of valid output tokens in the batch.
Therefore, group structure assigns role-specific relative credit, while all three roles jointly update the shared policy.
WebLoop is therefore sequential in rollout, role-specific in credit assignment, and joint in optimization.

\subsection{Executable Training Data Construction}
\label{sec:data_construction}

We construct executable Web generation tasks that pair natural-language specifications with browser-based functional supervision for WebLoop training.

\nbf{Task scope.}
The corpus covers diverse Web applications, including e-commerce, office productivity, education, developer tools, and data analytics.
Across these domains, we instantiate functional scenarios involving core operations, management functions, user settings, data presentation, and multi-step workflows.

\nbf{Construction pipeline.}
For each scenario, we construct explicit requirements that specify functions and interaction entry points, together with implicit requirements covering behavioral constraints such as state consistency, operation ordering, boundary handling, and reset behavior.
Motivated by WebRISE, we convert these requirements into an initial-state contract and an executable Interaction Contract Graph (ICG) containing browser interactions and functional assertions.
We then generate a reference implementation, execute the associated tests in a real browser, and retain tasks with transition validity $T\geq0.60$.
Further construction details are provided in~\cref{app:data}.

\nbf{Dataset statistics.}
The resulting corpus contains 2,880 tasks across 21 domains and 1,442 scenarios, with 11.87 requirements per task on average: 5.47 explicit and 6.39 implicit, with implicit requirements accounting for 53.9\%.
The ICGs average 10.64 states, 11.40 transitions, and 28.15 executable assertions, enabling fine-grained functional supervision.

\section{Experiments}
\label{sec:experiments}

\subsection{Experimental Setup}
\label{sec:experimental_setup}

\nbf{Benchmarks and metrics.}
We evaluate functional correctness on WebRise~\citep{meng2026webrise} and WebGen-Bench~\citep{lu2025webgenbench}.
WebRise contains 442 tasks and reports transition validity $T$, explicit requirement coverage $R_{\mathrm e}$, implicit requirement coverage $R_{\mathrm i}$, and their average Overall score.
For consistent comparison, all controlled training uses Text-conditioned tasks, and the main WebRise results are evaluated under the Text modality.
WebGen-Bench contains 101 website tasks with 647 functional test items, and we use the Bolt.diy framework with automatic starter-template selection disabled to generate websites from an empty workspace.
Following the official protocol, we report Yes, Partial, No, Start-Fail, and Accuracy. ``N/A'' indicates that model outputs could not be parsed by the harness due to incompatible output formatting.

\nbf{Baselines.}
We compare against representative general-purpose models, including Qwen3.5-122B-A10B and Qwen3.5-397B-A17B~\citep{qwen35}, Kimi-K2.6~\citep{kimi_k26}, GLM-5.3~\citep{glm53}, Claude Opus 4.7~\citep{claude_opus47}, Gemini 3.1 Pro~\citep{gemini31pro}, and GPT-6 Astra~\citep{gpt6astra}.
We also include Web-generation specialized models, including WebGen-LM~\citep{lu2025webgenbench}, WebGen-Agent~\citep{lu2025webgenagent}, and UIGEN-T2~\citep{uigen_t2}.
For controlled same-backbone comparisons, we evaluate the original Qwen3.5 model, Generate--Critique--Refine without RL, and alternative critique-learning methods Critique-GRPO~\citep{critique_grpo} and Critique-RL~\citep{Critique-RL}.
Throughout the paper, ``Gen'' denotes the first-pass code produced by the Generator of the trained checkpoint, while ``Refine'' denotes the revised code after one complete Generator--Critic--Refiner loop from the same checkpoint.

\nbf{Models and training.}
We instantiate WebLoop with Qwen3.5-9B and Qwen3.5-27B, with all controlled ablations conducted on the 9B model.
Training uses 3K text-conditioned Web generation tasks with a task batch size of 64.
For each task, we sample three initial generations, three critiques for the selected imperfect draft, and three refinements per critique, yielding 12 executable Web candidates per rollout.
The Critic is trained with discriminability alone for the first 15 updates and then with the combined discriminability and helpfulness reward using $\lambda=0.2$.
Unless otherwise stated, Generator, Critic, and Refiner samples are jointly optimized with GRPO.
Further training, rollout, and optimization details are provided in~\cref{app:implementation}.

\subsection{Main Results}
\label{sec:main_results}

\begin{table}[!t]
\caption{
\textbf{Main results on WebRise and WebGen-Bench.}
General-purpose and Web-generation specialized models are included as capability references, while the Qwen3.5 block provides controlled comparisons.
\textbf{Bold} denotes the best result within each reference group or controlled backbone size according to the metric direction.
$^\dagger$ denotes proprietary models.
}
\label{tab:mainresult}
\begin{center}
\small
\setlength{\tabcolsep}{4pt}
\renewcommand{\arraystretch}{1.08}
\resizebox{\textwidth}{!}{
\begin{tabular}{l|cccc|ccccc}
\toprule
\multirow{2}{*}{\textbf{Model / Method}}
& \multicolumn{4}{c|}{\textbf{WebRise}}
& \multicolumn{5}{c}{\textbf{WebGen-Bench}} \\
\cmidrule(lr){2-5}
\cmidrule(lr){6-10}
& $T$ $\uparrow$
& $R_e$ $\uparrow$
& $R_i$ $\uparrow$
& \textbf{Overall} $\uparrow$
& Yes\,(\%) $\uparrow$
& Partial\,(\%) $\uparrow$
& No\,(\%) $\downarrow$
& Start-Fail\,(\%) $\downarrow$
& \textbf{Acc.} $\uparrow$ \\
\midrule

\multicolumn{10}{l}{\textit{General-Purpose Models (Open-Source \& Proprietary)}} \\
\midrule
Qwen3.5-122B-A10B
& 38.0 & 48.9 & 35.2 & 40.7
& 9.9 & 6.2 & 78.1 & 5.9 & 13.0 \\

Kimi-K2.6
& 44.6 & 54.2 & 41.8 & 46.9
& 25.7 & 11.1 & 55.3 & 7.9 & 31.2 \\

Qwen3.5-397B-A17B
& 45.7 & 57.2 & 42.8 & 48.6
& 16.1 & 9.1 & 70.0 & 4.8 & 20.6 \\

GLM-5.3
& 59.5 & 77.1 & 54.9 & 63.8
& 20.6 & 11.9 & 67.5 & \textbf{0.0} & 26.5 \\

Claude Opus 4.7$^\dagger$
& 48.8 & 57.6 & 45.8 & 50.7
& \textbf{47.4} & \textbf{15.8} & \textbf{35.5} & 1.2 & \textbf{55.3} \\

Gemini 3.1 Pro$^\dagger$
& 50.7 & 61.1 & 47.5 & 53.1
& 40.2 & 12.5 & 47.0 & 0.3 & 46.4 \\

GPT-6 Astra$^\dagger$
& \textbf{62.5} & \textbf{78.9} & \textbf{57.7} & \textbf{66.4}
& 37.2 & 11.9 & 49.5 & 1.4 & 43.2 \\

\midrule
\multicolumn{10}{l}{\textit{Web-Generation Specialized Models}} \\
\midrule

WebGen-LM-7B
& 5.5 & 17.7 & 6.1 & 9.8
& 2.8 & 2.0 & 94.1 & 1.1 & 3.8 \\

WebGen-Agent-7B (Step-GRPO)
& 5.8 & 23.7 & 6.9 & 12.1
& 1.9 & \textbf{7.1} & \textbf{90.6} & 0.5 & \textbf{5.4} \\

WebGen-LM-32B
& 6.0 & 13.4 & 5.4 & 8.3
& 2.6 & 1.4 & 95.8 & \textbf{0.2} & 3.3 \\

UIGEN-T2-7B
& 12.1 & 33.0 & 11.1 & 18.7
& N/A & N/A & N/A & N/A & N/A \\

WebGen-LM-14B
& \textbf{15.3} & \textbf{35.5} & \textbf{14.5} & \textbf{21.8}
& \textbf{2.9} & 1.1 & 94.3 & 1.7 & 3.5 \\

\midrule
\multicolumn{10}{l}{\textit{Controlled Comparisons: Qwen3.5}} \\
\midrule

Base (Qwen3.5-9B)
& 23.2 & 45.3 & 22.1 & 30.2
& 18.4 & 10.2 & 67.1 & 4.3 & 23.5 \\

\quad Generate--Critique--Refine (w/o RL)
& 25.8 & 47.9 & 24.6 & 32.8
& 26.3 & 9.6 & 59.2 & 4.9 & 31.1 \\

\quad Critique-GRPO
& 27.6 & 50.0 & 26.2 & 34.6
& 13.4 & 8.3 & 76.7 & \textbf{1.5} & 17.6 \\

\quad Critique-RL
& 26.9 & 48.8 & 25.6 & 33.8
& 24.1 & 8.8 & 62.0 & 5.1 & 28.5 \\

\quad \textbf{WebLoop} (Gen)
& 31.7 & 55.4 & 30.0 & 39.1
& 28.7 & 10.8 & 56.9 & 3.6 & 34.2 \\

\rowcolor{oursgray}
\quad \textbf{WebLoop} (Refine)
& \textbf{34.6} & \textbf{57.9} & \textbf{32.0} & \textbf{41.5}
& \textbf{31.7} & \textbf{14.4} & \textbf{48.4} & 5.6 & \textbf{38.9} \\

\cmidrule(lr){1-10}

Base (Qwen3.5-27B)
& 36.3 & 47.3 & 34.3 & 39.3
& 24.3 & 11.0 & 62.1 & 2.6 & 29.8 \\

\quad Generate--Critique--Refine (w/o RL)
& 40.5 & 63.1 & 37.6 & 47.1
& 32.6 & 11.0 & 53.8 & 2.6 & 38.1 \\

\quad \textbf{WebLoop} (Gen)
& 43.8 & 66.6 & 40.1 & 50.2
& 26.0 & 13.0 & 60.9 & \textbf{0.2} & 32.5 \\

\rowcolor{oursgray}
\quad \textbf{WebLoop} (Refine)
& \textbf{46.1} & \textbf{68.0} & \textbf{42.7} & \textbf{52.2}
& \textbf{36.0} & \textbf{13.6} & \textbf{50.2} & \textbf{0.2} & \textbf{42.8} \\

\bottomrule
\end{tabular}
}
\end{center}
\end{table}

\nbf{Functional Web generation.}
\cref{tab:mainresult} shows that WebLoop consistently improves functional correctness across both backbone sizes and benchmarks.
With Qwen3.5-9B, WebLoop (Refine) reaches 41.5 Overall on WebRise and 38.9\% accuracy on WebGen-Bench, improving the base model by 11.3 and 15.4 points and surpassing Qwen3.5-122B-A10B on both benchmarks.
At 27B, WebLoop further reaches 52.2 and 42.8\%, outperforming Qwen3.5-397B-A17B and the 1T-parameter Kimi-K2.6 despite a substantially smaller total parameter footprint.
It also approaches proprietary frontier models, trailing Gemini 3.1 Pro by only 0.9 points on WebRise and 3.6 points on WebGen-Bench, and GPT-6 Astra by 0.4 points on WebGen-Bench.

\nbf{First-pass generation and model scaling.}
As shown in~\cref{tab:mainresult}, WebLoop improves both first-pass generation and final refinement.
WebLoop (Gen) raises Qwen3.5-9B from 30.2 to 39.1 on WebRise and from 23.5\% to 34.2\% on WebGen-Bench.
These gains indicate that loop-level training strengthens the underlying generation policy itself, rather than benefiting only critique-conditioned revision.
At 27B, the corresponding gains are 39.3 to 50.2 and 29.8\% to 32.5\%, respectively.
The complete loop further reaches 52.2 and 42.8\%, showing that the benefits persist with model scaling and continue to compound through refinement.

\begin{table*}[t]
\caption{
\textbf{Cross-modal generalization on WebRise.}
WebLoop is trained only on Text-conditioned tasks and evaluated on all five input modalities using the same checkpoint.
\textbf{Avg.} averages the modality-level $(T+R_e+R_i)/3$ scores.
}
\label{tab:abl_modality}
\centering
\scriptsize
\setlength{\tabcolsep}{2.4pt}
\renewcommand{\arraystretch}{1.08}
\resizebox{\textwidth}{!}{
\begin{tabular}{l|ccc|ccc|ccc|ccc|ccc|c}
\toprule
\multirow{2}{*}{\textbf{Model / Stage}}
& \multicolumn{3}{c|}{\textbf{Text}}
& \multicolumn{3}{c|}{\textbf{Markdown}}
& \multicolumn{3}{c|}{\textbf{Sketch}}
& \multicolumn{3}{c|}{\textbf{Image}}
& \multicolumn{3}{c|}{\textbf{Video}}
& \multirow{2}{*}{\textbf{Avg.} $\uparrow$} \\
\cmidrule(lr){2-4}
\cmidrule(lr){5-7}
\cmidrule(lr){8-10}
\cmidrule(lr){11-13}
\cmidrule(lr){14-16}
& $T$ $\uparrow$ & $R_e$ $\uparrow$ & $R_i$ $\uparrow$
& $T$ $\uparrow$ & $R_e$ $\uparrow$ & $R_i$ $\uparrow$
& $T$ $\uparrow$ & $R_e$ $\uparrow$ & $R_i$ $\uparrow$
& $T$ $\uparrow$ & $R_e$ $\uparrow$ & $R_i$ $\uparrow$
& $T$ $\uparrow$ & $R_e$ $\uparrow$ & $R_i$ $\uparrow$
& \\
\midrule
Base (Qwen3.5-9B)
& 23.2 & 45.3 & 22.1
& 28.2 & 51.1 & 26.0
& 25.2 & 48.8 & 21.9
& 23.0 & 46.2 & 22.8
& 24.7 & 45.7 & 23.9
& 31.9 \\
\quad \textbf{WebLoop} (Gen)
& 31.7 & 55.4 & 30.0
& 30.5 & 55.0 & 28.4
& 28.7 & 53.4 & 26.2
& 27.7 & 51.9 & 26.1
& 31.8 & 54.1 & 30.9
& 37.5 \\
\quad \textbf{WebLoop} (Refine)
& \textbf{34.6} & \textbf{57.9} & \textbf{32.0}
& \textbf{34.4} & \textbf{57.5} & \textbf{32.1}
& \textbf{31.1} & \textbf{55.0} & \textbf{28.1}
& \textbf{30.6} & \textbf{53.1} & \textbf{29.0}
& \textbf{35.9} & \textbf{57.3} & \textbf{33.6}
& \textbf{40.1} \\
\bottomrule
\end{tabular}
}
\end{table*}

\nbf{Generalization across input modalities.}
\cref{tab:abl_modality} evaluates whether Text-only WebLoop training transfers to other input modalities.
The average WebRise score improves from 31.9 for the base model to 37.5 for WebLoop (Gen) and 40.1 after the complete loop, with consistent gains across Text, Markdown, Sketch, Image, and Video.
The improvements also span all three functional metrics, demonstrating that the learned generation and refinement capabilities generalize beyond the text-conditioned training distribution.

\section{Analysis and Discussion}

\begin{table}[!t]
\caption{
\textbf{Where does the gain come from?}
Controlled comparison of training schemes on WebRise and WebGen-Bench.
All variants use the same Qwen3.5-9B backbone, training data, and rollout budget, and differ only in which roles are optimized and whether they are trained jointly or sequentially.
\textbf{Bold} denotes the best result according to the metric direction.
}
\label{tab:abl_mechanism}
\begin{center}
\small
\setlength{\tabcolsep}{4pt}
\renewcommand{\arraystretch}{1.12}
\resizebox{\textwidth}{!}{
\begin{tabular}{l|l|cccc|ccccc}
\toprule
\multirow{2}{*}{\textbf{Training Scheme / Output}}
& \multirow{2}{*}{\textbf{Optimized Roles}}
& \multicolumn{4}{c|}{\textbf{WebRise}}
& \multicolumn{5}{c}{\textbf{WebGen-Bench}} \\
\cmidrule(lr){3-6}
\cmidrule(lr){7-11}
& & $T$ $\uparrow$
& $R_e$ $\uparrow$
& $R_i$ $\uparrow$
& \textbf{Overall} $\uparrow$
& Yes\,(\%) $\uparrow$
& Partial\,(\%) $\uparrow$
& No\,(\%) $\downarrow$
& Start-Fail\,(\%) $\downarrow$
& \textbf{Acc.} $\uparrow$ \\
\midrule

Base
& --
& 23.2 & 45.3 & 22.1 & 30.2
& 18.4 & 10.2 & 67.1 & 4.3 & 23.5 \\

\midrule
Generation Only (Gen)
& Gen
& 29.5 & 54.1 & 27.3 & 36.9
& 17.8 & 10.5 & 67.9 & 3.9 & 23.0 \\

\midrule
Refine w/o Critique (Gen)
& \multirow{2}{*}{Gen + Refine}
& 28.2 & 53.3 & 25.2 & 35.6
& 16.7 & 11.9 & 70.5 & 0.9 & 22.6 \\

Refine w/o Critique (Refine)
&
& 32.9 & 57.3 & 29.7 & 40.0
& 21.6 & 10.0 & 68.3 & \textbf{0.0} & 26.7 \\

\midrule
Staged Training (Gen)
& \multirow{2}{*}{Gen $\rightarrow$ Critic + Refine}
& 27.6 & 52.0 & 25.9 & 35.2
& 20.7 & 11.1 & 66.0 & 2.2 & 26.3 \\

Staged Training (Refine)
&
& 32.9 & 56.4 & 30.9 & 40.1
& 29.8 & 12.5 & 55.5 & 2.2 & 36.1 \\

\midrule
WebLoop (Gen)
& \multirow{2}{*}{Gen + Critic + Refine}
& 31.7 & 55.4 & 30.0 & 39.1
& 28.7 & 10.8 & 56.9 & 3.6 & 34.2 \\

WebLoop (Refine)
&
& \textbf{34.6}
& \textbf{57.9}
& \textbf{32.0}
& \textbf{41.5}
& \textbf{31.7}
& \textbf{14.4}
& \textbf{48.4}
& 5.6
& \textbf{38.9} \\

\bottomrule
\end{tabular}
}
\end{center}
\end{table}

\subsection{Where Does the Gain Come From?}
\label{sec:gain_analysis}

\nbf{Beyond additional refinement.}
\Cref{tab:abl_mechanism} shows that WebLoop's gains cannot be explained by an extra refinement pass alone.
Generation Only reaches 36.9 on WebRise and 23.0\% on WebGen-Bench, while Refine w/o Critique improves the final output to 40.0 and 26.7\%, respectively.
WebLoop further reaches 41.5 and 38.9\%, indicating that learned critique provides additional value beyond second-pass generation.
Joint optimization is also important: Staged Training reaches 40.1 and 36.1\% after refinement, while WebLoop achieves 41.5 and 38.9\%, with a larger gap already visible in first-pass generation.
These results show that WebLoop benefits from jointly learning generation, critique, and refinement rather than treating refinement or Critic learning as isolated stages.

\begin{table}[!t]
\caption{
\textbf{Critic reward ablation.}
All variants use the same Qwen3.5-9B backbone, training data, and rollout structure and differ only in the Critic reward.
$D$ denotes discriminability and $H$ denotes downstream helpfulness.
\textbf{Bold} denotes the best result according to the metric direction.
}
\label{tab:abl_critique}
\begin{center}
\small
\setlength{\tabcolsep}{4pt}
\renewcommand{\arraystretch}{1.12}
\resizebox{\textwidth}{!}{
\begin{tabular}{l|l|cccc|ccccc}
\toprule
\multirow{2}{*}{\textbf{Critic Training / Output}}
& \multirow{2}{*}{$\mathbf{R_{\mathrm{crit}}}$}
& \multicolumn{4}{c|}{\textbf{WebRise}}
& \multicolumn{5}{c}{\textbf{WebGen-Bench}} \\
\cmidrule(lr){3-6}
\cmidrule(lr){7-11}
& & $T$ $\uparrow$
& $R_e$ $\uparrow$
& $R_i$ $\uparrow$
& \textbf{Overall} $\uparrow$
& Yes\,(\%) $\uparrow$
& Partial\,(\%) $\uparrow$
& No\,(\%) $\downarrow$
& Start-Fail\,(\%) $\downarrow$
& \textbf{Acc.} $\uparrow$ \\
\midrule

Base
& --
& 23.2 & 45.3 & 22.1 & 30.2
& 18.4 & 10.2 & 67.1 & 4.3 & 23.5 \\

\midrule
Discriminability Only (Gen)
& \multirow{2}{*}{$D$}
& 30.0 & 53.6 & 27.2 & 36.9
& 22.3 & 8.7 & 65.8 & 3.2 & 26.6 \\

Discriminability Only (Refine)
&
& 33.4 & 57.3 & 30.7 & 40.5
& 27.8 & 10.7 & 59.4 & \textbf{2.2} & 33.2 \\

\midrule
Helpfulness Only (Gen)
& \multirow{2}{*}{$H$}
& 27.6 & 50.7 & 24.7 & 34.3
& 18.2 & 9.0 & 68.2 & 4.6 & 22.7 \\

Helpfulness Only (Refine)
&
& 31.3 & 54.1 & 28.7 & 38.0
& 28.1 & 13.0 & 56.3 & 2.6 & 34.6 \\

\midrule
Mixed from Start (Gen)
& \multirow{2}{*}{$H+\lambda D$}
& 30.2 & 54.0 & 27.8 & 37.3
& 24.4 & 9.3 & 62.9 & 3.4 & 29.1 \\

Mixed from Start (Refine)
&
& 32.7 & 56.5 & 29.4 & 39.5
& 28.0 & 10.5 & 58.3 & 3.2 & 33.2 \\

\midrule
\textbf{WebLoop} (Gen)
& \multirow{2}{*}{$D \rightarrow H+\lambda D$}
& 31.7 & 55.4 & 30.0 & 39.1
& 28.7 & 10.8 & 56.9 & 3.6 & 34.2 \\

\textbf{WebLoop} (Refine)
&
& \textbf{34.6}
& \textbf{57.9}
& \textbf{32.0}
& \textbf{41.5}
& \textbf{31.7}
& \textbf{14.4}
& \textbf{48.4}
& 5.6
& \textbf{38.9} \\

\bottomrule
\end{tabular}
}
\end{center}
\end{table}

\begin{figure}[!htb]
\centering
\includegraphics[width=\linewidth]{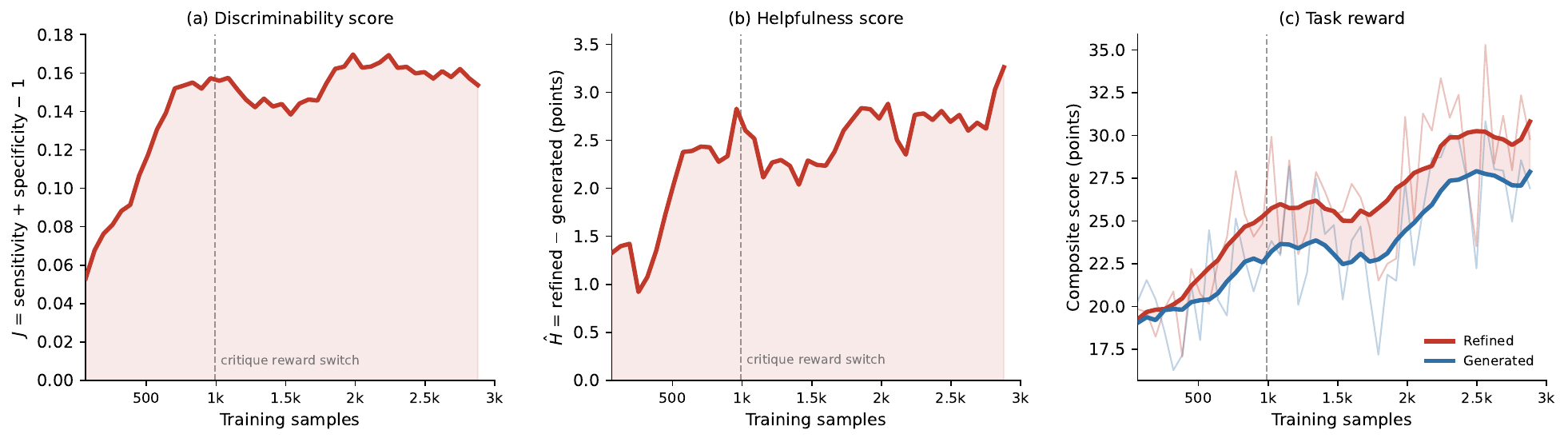}
\caption{
\textbf{Critic learning dynamics.}
(a) Discriminability $D$.
(b) Downstream helpfulness $H$.
(c) Generator and Refiner execution rewards.
Thin and thick curves denote per-step values and moving averages; the dashed line marks the reward switch after 15 updates (around 1K samples).
}
\label{fig:training_dynamics}
\end{figure}

\subsection{What Makes a Useful Critique?}
\label{sec:critic_analysis}

\nbf{Complementary critic signals and staged learning.}
\Cref{tab:abl_critique} shows that discriminability and helpfulness capture complementary aspects of critique quality.
Optimizing either signal alone underperforms WebLoop, and their relative ranking differs across benchmarks: discriminability performs better on WebRise, while helpfulness is stronger on WebGen-Bench.
Moreover, mixing the two signals from the beginning still falls short of the staged objective, indicating that both the reward composition and its scheduling matter.
WebLoop reaches 41.5 on WebRise and 38.9\% on WebGen-Bench, supporting a curriculum that first establishes reliable diagnosis and then introduces downstream utility.
This allows the Critic to learn feedback that is both correct and actionable.

\nbf{Critic learning dynamics.}
\Cref{fig:training_dynamics} tracks discriminability $D$, downstream helpfulness $H$, and executable reward throughout training.
We use the first 15 updates as a diagnostic warm-up, optimizing only $D$ so that the Critic first learns reliable requirement-level judgments from direct execution labels.
We then switch to $H+\lambda D$, introducing the delayed downstream signal while retaining diagnostic supervision.
This staged schedule avoids relying on refinement-dependent utility before the Critic has learned a stable notion of correctness.
Refined outputs remain consistently stronger than first-pass generations, indicating that improved Critic behavior translates into executable gains.
Additional training-dynamics and co-adaptation analyses are provided in~\cref{app:training_dynamics}.


\section{Conclusion}

We presented \textbf{WebLoop}, an execution-grounded framework that jointly learns generation, critique, and refinement for functional Web generation.
WebLoop treats critique as a learnable intermediate decision, assigning execution-derived credit through complementary discriminability and downstream helpfulness signals while jointly optimizing all three roles within a shared policy.
Across WebRise and WebGen-Bench, WebLoop improves both first-pass generation and final refinement, with gains that persist across model scales and transfer from text-only training to multimodal inputs.
Controlled ablations show that these gains cannot be explained by an additional refinement pass alone, suggesting that executable supervision can be used not only to optimize final artifacts, but also to learn how they should be diagnosed and improved.

\clearpage

\subsection*{AI Use Statement}

Generative AI tools, including ChatGPT and AI coding assistants such as Codex, were used during the preparation of this work.
They assisted with language editing, improving clarity and organization, and the implementation, debugging, and refinement of selected code components.
AI models were also used to assist the generation and refinement of portions of the executable training data under predefined data-construction protocols.
AI-assisted training data were subsequently filtered and validated through the authors' specified quality-control and browser-execution procedures.
The authors made and verified the final decisions regarding the research questions, methodological design, experimental protocol, analysis, and conclusions.
Generative AI was not used to fabricate, alter, or post hoc select reported experimental results; all reported metrics were computed from actual model outputs using the stated evaluation pipelines.
The authors reviewed all AI-assisted content and take full responsibility for the final manuscript, implementation, data, and reported findings.

\subsection*{Reproducibility Statement}

We provide the formulation, training procedure, data construction pipeline, and evaluation protocol required to reproduce WebLoop.
The main paper describes the Generator--Critic--Refiner loop, execution-grounded Critic credit assignment, joint group-relative optimization, and experimental protocol.
\Cref{app:implementation} provides training and rollout configurations, complete objective definitions, browser evaluation details, and controlled-baseline adaptations, while \cref{app:data} details executable task construction and filtering.
The complete role-specific prompts are provided in~\cref{app:prompts}.
Unless otherwise stated, controlled comparisons use the same backbone initialization, training-task distribution, browser evaluation infrastructure, and metric computation.
The core code is provided in the supplementary material accompanying this paper.
Source code, configurations, and data-construction assets will be publicly released upon acceptance to facilitate reproducibility and future research.


\bibliography{iclr2027_conference}

@inproceedings{lu2025webgenbench,
  author       = {Zimu Lu and
                  Yunqiao Yang and
                  Houxing Ren and
                  Haotian Hou and
                  Han Xiao and
                  Ke Wang and
                  Weikang Shi and
                  Aojun Zhou and
                  Mingjie Zhan and
                  Hongsheng Li},
  title        = {WebGen-Bench: Evaluating LLMs on Generating Interactive and Functional
                  Websites from Scratch},
  booktitle    = {NeurIPS},
  year         = {2025}
}

@article{meng2026webrise,
  author       = {Yuxin Meng and
                  Yuhan Suo and
                  Junjie Wang and
                  Yuhan Sun and
                  Yiyao Yu and
                  Ruixu Zhang and
                  Ruining Hu and
                  Yubin Wang and
                  Shouwei Ruan and
                  Bin Wang and
                  Yuxiang Zhang and
                  Yujiu Yang},
  title        = {WebRISE: Requirement-Induced State Evaluation for MLLM-Generated Web
                  Artifacts},
  journal      = {CoRR},
  volume       = {abs/2606.03220},
  year         = {2026}
}

@article{jiang2026webgenr1,
  author       = {Juyong Jiang and
                  Chenglin Cai and
                  Chansung Park and
                  Jiasi Shen and
                  Sung Hun Kim and
                  Jianguo Li and
                  Yue Wang},
  title        = {WebGen-R1: Incentivizing Large Language Models to Generate Functional
                  and Aesthetic Websites with Reinforcement Learning},
  journal      = {CoRR},
  volume       = {abs/2604.20398},
  year         = {2026}
}

@article{chen2026webgrader,
  author       = {Boshui Chen and
                  Huiping Liu and
                  Shaolei Zhang},
  title        = {WebGrader: Training LLMs for Web Development with Self-Evolving Programmatic
                  Grader},
  journal      = {CoRR},
  volume       = {abs/2608.06474},
  year         = {2026}
}

@inproceedings{madaan2023selfrefine,
  author       = {Aman Madaan and
                  Niket Tandon and
                  Prakhar Gupta and
                  Skyler Hallinan and
                  Luyu Gao and
                  Sarah Wiegreffe and
                  Uri Alon and
                  Nouha Dziri and
                  Shrimai Prabhumoye and
                  Yiming Yang and
                  Shashank Gupta and
                  Bodhisattwa Prasad Majumder and
                  Katherine Hermann and
                  Sean Welleck and
                  Amir Yazdanbakhsh and
                  Peter Clark},
  title        = {Self-Refine: Iterative Refinement with Self-Feedback},
  booktitle    = {NeurIPS},
  year         = {2023}
}

@inproceedings{shinn2023reflexion,
  author       = {Noah Shinn and
                  Federico Cassano and
                  Ashwin Gopinath and
                  Karthik Narasimhan and
                  Shunyu Yao},
  title        = {Reflexion: language agents with verbal reinforcement learning},
  booktitle    = {NeurIPS},
  year         = {2023}
}

@inproceedings{gou2023critic,
  author       = {Zhibin Gou and
                  Zhihong Shao and
                  Yeyun Gong and
                  Yelong Shen and
                  Yujiu Yang and
                  Nan Duan and
                  Weizhu Chen},
  title        = {{CRITIC:} Large Language Models Can Self-Correct with Tool-Interactive
                  Critiquing},
  booktitle    = {{ICLR}},
  publisher    = {OpenReview.net},
  year         = {2024}
}

@inproceedings{chen2024selfdebug,
  author       = {Xinyun Chen and
                  Maxwell Lin and
                  Nathanael Sch{\"{a}}rli and
                  Denny Zhou},
  title        = {Teaching Large Language Models to Self-Debug},
  booktitle    = {{ICLR}},
  publisher    = {OpenReview.net},
  year         = {2024}
}

@article{lu2025webgenagent,
  author       = {Zimu Lu and
                  Houxing Ren and
                  Yunqiao Yang and
                  Ke Wang and
                  Zhuofan Zong and
                  Junting Pan and
                  Mingjie Zhan and
                  Hongsheng Li},
  title        = {WebGen-Agent: Enhancing Interactive Website Generation with Multi-Level
                  Feedback and Step-Level Reinforcement Learning},
  journal      = {CoRR},
  volume       = {abs/2509.22644},
  year         = {2025}
}

@inproceedings{li2026relook,
  author       = {Yuhang Li and
                  Chenchen Zhang and
                  Ruilin Lv and
                  Ao Liu and
                  Ken Deng and
                  Yuanxing Zhang and
                  Jiaheng Liu and
                  Bo Zhou},
  title        = {ReLook: Vision-Grounded {RL} with a Multimodal {LLM} Critic for Agentic
                  Web Coding},
  booktitle    = {{ACL} {(1)}},
  pages        = {25471--25485},
  publisher    = {Association for Computational Linguistics},
  year         = {2026}
}

@article{Critique-RL,
  author       = {Zhiheng Xi and
                  Jixuan Huang and
                  Xin Guo and
                  Boyang Hong and
                  Dingwen Yang and
                  Xiaoran Fan and
                  Shuo Li and
                  Zehui Chen and
                  Junjie Ye and
                  Siyu Yuan and
                  Zhengyin Du and
                  Xuesong Yao and
                  Yufei Xu and
                  Jiecao Chen and
                  Rui Zheng and
                  Tao Gui and
                  Qi Zhang and
                  Xuanjing Huang},
  title        = {{Critique-RL}: Training Language Models for Critiquing through Two-Stage
                  Reinforcement Learning},
  journal      = {CoRR},
  volume       = {abs/2510.24320},
  year         = {2025}
}

@article{websight,
  author       = {Hugo Lauren{\c{c}}on and
                  L{\'{e}}o Tronchon and
                  Victor Sanh},
  title        = {Unlocking the conversion of Web Screenshots into {HTML} Code with
                  the WebSight Dataset},
  journal      = {CoRR},
  volume       = {abs/2403.09029},
  year         = {2024}
}

@inproceedings{design2code,
  author       = {Chenglei Si and
                  Yanzhe Zhang and
                  Ryan Li and
                  Zhengyuan Yang and
                  Ruibo Liu and
                  Diyi Yang},
  title        = {Design2Code: Benchmarking Multimodal Code Generation for Automated
                  Front-End Engineering},
  booktitle    = {{NAACL} (Long Papers)},
  pages        = {3956--3974},
  publisher    = {Association for Computational Linguistics},
  year         = {2025}
}

@inproceedings{web2code,
  author       = {Sukmin Yun and
                  Haokun Lin and
                  Rusiru Thushara and
                  Mohammad Qazim Bhat and
                  Yongxin Wang and
                  Zutao Jiang and
                  Mingkai Deng and
                  Jinhong Wang and
                  Tianhua Tao and
                  Junbo Li and
                  Haonan Li and
                  Preslav Nakov and
                  Timothy Baldwin and
                  Zhengzhong Liu and
                  Eric P. Xing and
                  Xiaodan Liang and
                  Zhiqiang Shen},
  title        = {Web2Code: {A} Large-scale Webpage-to-Code Dataset and Evaluation Framework
                  for Multimodal LLMs},
  booktitle    = {NeurIPS},
  year         = {2024}
}

@article{frontendbench,
  author       = {Hongda Zhu and
                  Yiwen Zhang and
                  Bing Zhao and
                  Jingzhe Ding and
                  Siyao Liu and
                  Tong Liu and
                  Dandan Wang and
                  Yanan Liu and
                  Zhaojian Li},
  title        = {FrontendBench: {A} Benchmark for Evaluating LLMs on Front-End Development
                  via Automatic Evaluation},
  journal      = {CoRR},
  volume       = {abs/2506.13832},
  year         = {2025}
}

@inproceedings{selfcorrect_limit,
  author       = {Jie Huang and
                  Xinyun Chen and
                  Swaroop Mishra and
                  Huaixiu Steven Zheng and
                  Adams Wei Yu and
                  Xinying Song and
                  Denny Zhou},
  title        = {Large Language Models Cannot Self-Correct Reasoning Yet},
  booktitle    = {{ICLR}},
  publisher    = {OpenReview.net},
  year         = {2024}
}

@inproceedings{score,
  author       = {Aviral Kumar and
                  Vincent Zhuang and
                  Rishabh Agarwal and
                  Yi Su and
                  John D. Co{-}Reyes and
                  Avi Singh and
                  Kate Baumli and
                  Shariq Iqbal and
                  Colton Bishop and
                  Rebecca Roelofs and
                  Lei M. Zhang and
                  Kay McKinney and
                  Disha Shrivastava and
                  Cosmin Paduraru and
                  George Tucker and
                  Doina Precup and
                  Feryal M. P. Behbahani and
                  Aleksandra Faust},
  title        = {Training Language Models to Self-Correct via Reinforcement Learning},
  booktitle    = {{ICLR}},
  publisher    = {OpenReview.net},
  year         = {2025}
}

@article{criticgpt,
  author       = {Nat McAleese and
                  Rai Michael Pokorny and
                  Juan Felipe Ceron Uribe and
                  Evgenia Nitishinskaya and
                  Maja Trebacz and
                  Jan Leike},
  title        = {{LLM} Critics Help Catch {LLM} Bugs},
  journal      = {CoRR},
  volume       = {abs/2407.00215},
  year         = {2024}
}

@article{selftaught,
  author       = {Tianlu Wang and
                  Ilia Kulikov and
                  Olga Golovneva and
                  Ping Yu and
                  Weizhe Yuan and
                  Jane Dwivedi{-}Yu and
                  Richard Yuanzhe Pang and
                  Maryam Fazel{-}Zarandi and
                  Jason Weston and
                  Xian Li},
  title        = {Self-Taught Evaluators},
  journal      = {CoRR},
  volume       = {abs/2408.02666},
  year         = {2024}
}

@inproceedings{rco,
  author       = {Tianshu Yu and
                  Chao Xiang and
                  Mingchuan Yang and
                  Pei Ke and
                  Bosi Wen and
                  Cunxiang Wang and
                  Jiale Cheng and
                  Li Zhang and
                  Xinyu Mu and
                  Chuxiong Sun and
                  Minlie Huang},
  title        = {Training Language Model to Critique for Better Refinement},
  booktitle    = {{ACL} (Findings)},
  series       = {Findings of {ACL}},
  volume       = {{ACL} 2025},
  pages        = {26760--26804},
  publisher    = {Association for Computational Linguistics},
  year         = {2025}
}

@misc{qwen35,
  title  = {{Qwen3.5}: Towards Native Multimodal Agents},
  author = {{Qwen Team}},
  month  = {February},
  year   = {2026},
  url    = {https://qwen.ai/blog?id=qwen3.5}
}

@misc{kimi_k26,
  title        = {{Kimi K2.6}},
  author       = {{Moonshot AI}},
  year         = {2026},
  month        = {April},
  howpublished = {Model Card},
  url          = {https://huggingface.co/moonshotai/Kimi-K2.6}
}

@misc{glm53,
  title         = {{GLM-5}: From Vibe Coding to Agentic Engineering},
  author        = {{GLM-5 Team}},
  year          = {2026},
  eprint        = {2602.15763},
  archivePrefix = {arXiv},
  primaryClass  = {cs.LG},
  url           = {https://arxiv.org/abs/2602.15763}
}

@techreport{claude_opus47,
  title       = {{Claude Opus 4.7 System Card}},
  author      = {{Anthropic}},
  institution = {Anthropic},
  year        = {2026},
  month       = {April},
  url         = {https://www.anthropic.com/claude-opus-4-7-system-card}
}

@techreport{gemini31pro,
  title       = {{Gemini 3.1 Pro Model Card}},
  author      = {{Google DeepMind}},
  institution = {Google DeepMind},
  year        = {2026},
  month       = {February},
  url         = {https://storage.googleapis.com/deepmind-media/Model-Cards/Gemini-3-1-Pro-Model-Card.pdf}
}

@techreport{gpt6astra,
  title       = {{GPT-6 Astra System Card}},
  author      = {{OpenAI}},
  institution = {OpenAI},
  year        = {2026},
  month       = {September},
  url         = {https://deploymentsafety.openai.com/gpt-6-astra}
}

@article{critique_grpo,
  author       = {Xiaoying Zhang and
                  Hao Sun and
                  Yipeng Zhang and
                  Kaituo Feng and
                  Chaochao Lu and
                  Chao Yang and
                  Helen Meng},
  title        = {Critique-GRPO: Advancing {LLM} Reasoning with Natural Language and
                  Numerical Feedback},
  journal      = {CoRR},
  volume       = {abs/2506.03106},
  year         = {2025}
}

@misc{uigen_t2,
  title={UIGEN-T2: Scaling UI Generation with Reasoning on Qwen2.5-Coder-7B},
  author={tesslate},
  year={2024},
  publisher={Hugging Face},
  url={https://huggingface.co/tesslate/UIGEN-T2}
}
\bibliographystyle{iclr2027_conference}

\clearpage
\appendix
\section*{Appendix}

\appendix

\section{Implementation and Training Details}
\label{app:implementation}

We provide additional implementation details for WebLoop training, complete definitions of the objectives simplified in the main text, and the evaluation protocols used for controlled comparisons.

\subsection{Training and Rollout Configuration}
\label{app:training}

\nbf{Optimization setup.}
We instantiate WebLoop with Qwen3.5-9B and Qwen3.5-27B, and conduct all controlled ablations with Qwen3.5-9B.
Training uses the 2,880 executable tasks described in~\cref{sec:data_construction}, referred to as approximately 3K tasks in the main text, with a task batch size of 64.
We optimize the shared policy with GRPO using a learning rate of $1\times10^{-6}$ and no explicit KL penalty.
Rollouts use a temperature of $0.7$ and top-$p$ of $1.0$, with maximum prompt and response lengths of 16,384 tokens.

\nbf{Rollout structure.}
For each task, the Generator samples $n_g=3$ initial implementations.
After selecting one valid imperfect draft, the Critic samples $n_c=3$ critiques, each of which conditions $n_r=3$ independent refinements.
A complete rollout therefore contains 3 Generator outputs, 3 Critic outputs, and 9 Refiner outputs, including 12 executable Web pages.
If no valid imperfect draft exists, the task contributes only its Generator samples.
The three roles are sampled sequentially and jointly optimized in one GRPO update as described in~\cref{sec:joint_optimization}.

\nbf{Critic schedule.}
For the first 15 policy updates, the Critic is optimized only with diagnostic discriminability $D(c)$.
Subsequent updates use $H(c)+\lambda D(c)$ with $\lambda=0.2$, thereby introducing retrospective helpfulness while retaining requirement-level diagnostic supervision.
Generator, Critic, and Refiner share all model parameters throughout training.

\nbf{System implementation.}
Our implementation is built on \texttt{verl} with an FSDP2 actor and asynchronous vLLM rollout workers.
Policy weights are synchronized to the rollout workers after each update to maintain on-policy sampling.
Browser execution is parallelized independently from model rollout, with up to 72 concurrent evaluations and a per-example hard timeout of 2,400 seconds.

\subsection{Additional Objective Details}
\label{app:objective_details}

\nbf{Candidate selection.}
Let $\mathcal Y_x^{(1)}=\{y_i^{(1)}\}_{i=1}^{n_g}$ denote the initial implementations sampled for task $x$.
We retain candidates that receive a valid browser evaluation but do not fully solve the task:
\begin{equation}
\small
\mathcal Y_x^{-}
=
\left\{
y_i^{(1)}
\in
\mathcal Y_x^{(1)}
\mid
y_i^{(1)} \text{ is valid},
\;
r_{\mathrm{web}}\big(y_i^{(1)}\big)<1
\right\}.
\label{eq:app_failed_candidates}
\end{equation}
The shared base draft is the median-reward candidate in this set,
\begin{equation}
\small
y_b
=
\operatorname{Median}_{r_{\mathrm{web}}}
\left(
\mathcal Y_x^{-}
\right).
\label{eq:app_candidate_selection}
\end{equation}
This choice keeps all critiques anchored to the same implementation while avoiding both solved examples and unusually poor failures.
If $\mathcal Y_x^{-}$ is empty, no Critic or Refiner branch is expanded for that task.

\nbf{Complete discriminability definition.}
For the selected draft $y_b$, let $z_q\in\{0,1\}$ denote the execution-grounded status of requirement $r_q$, and let $\hat z_{j,q}$ be the corresponding prediction from critique $c_j$.
We define
\begin{equation}
\small
\mathcal F=\{q:z_q=0\},
\qquad
\mathcal P=\{q:z_q=1\},
\end{equation}
and the class-wise accuracies
\begin{equation}
\small
a_{\mathrm{fail}}(c_j)
=
\frac{1}{|\mathcal F|}
\sum_{q\in\mathcal F}
\mathbb I[\hat z_{j,q}=0],
\qquad
a_{\mathrm{pass}}(c_j)
=
\frac{1}{|\mathcal P|}
\sum_{q\in\mathcal P}
\mathbb I[\hat z_{j,q}=1],
\label{eq:app_class_accuracy}
\end{equation}
whenever the corresponding set is non-empty.
The full discriminability score is
\begin{equation}
\small
D(c_j)
=
\begin{cases}
a_{\mathrm{fail}}(c_j)+a_{\mathrm{pass}}(c_j)-1,
& |\mathcal F|>0,\ |\mathcal P|>0,\\
2a_{\mathrm{fail}}(c_j)-1,
& |\mathcal F|>0,\ |\mathcal P|=0,\\
2a_{\mathrm{pass}}(c_j)-1,
& |\mathcal F|=0,\ |\mathcal P|>0,\\
0,
& Q_x=0.
\end{cases}
\label{eq:app_diagnosis_score}
\end{equation}
When both classes are present, this reduces to Youden's $J$ and assigns equal importance to satisfied and violated requirements.
The single-class cases preserve the same $[-1,1]$ range.

\nbf{Expected refinement utility.}
The helpfulness of a critique is defined through the quality of refinements generated under that critique.
For a fixed $(x,y_b,c_j)$, the conditional refinement utility is
\begin{equation}
\small
U(c_j\mid x,y_b)
=
\mathbb E_{
y\sim
\pi_{\theta_{\mathrm{old}}}
(\cdot\mid p_{\mathrm{ref}}(x,y_b,c_j))
}
\left[
r_{\mathrm{web}}(y)
\right].
\label{eq:app_expected_utility}
\end{equation}
We estimate this expectation using the $n_r$ sampled refinements,
\begin{equation}
\small
\widehat U(c_j)
=
\frac{1}{n_r}
\sum_{k=1}^{n_r}
r_{\mathrm{web}}
\big(
y_{j,k}^{(2)}
\big),
\end{equation}
and define retrospective helpfulness as
\begin{equation}
\small
H(c_j)
=
\widehat U(c_j)
-
r_{\mathrm{web}}(y_b).
\label{eq:app_helpfulness}
\end{equation}
Thus, $H(c_j)$ measures the expected execution gain obtained when the current Refiner acts on the critique rather than the linguistic form of the critique itself.

\nbf{Role-specific rewards.}
For completeness, the scalar reward assigned to any policy output $o$ is
\begin{equation}
\small
r(o)
=
\begin{cases}
r_{\mathrm{web}}\big(y_i^{(1)}\big),
& o=y_i^{(1)},\\
r_{\mathrm{crit}}(c_j),
& o=c_j,\\
r_{\mathrm{web}}\big(y_{j,k}^{(2)}\big),
& o=y_{j,k}^{(2)}.
\end{cases}
\label{eq:app_role_rewards}
\end{equation}
These heterogeneous rewards are normalized only within the role-specific comparison groups defined in~\cref{sec:joint_optimization}.
Critiques that cannot be parsed into the required requirement-level judgments receive the minimum Critic reward and do not instantiate downstream refinement branches.

\subsection{Browser Evaluation and Baseline Adaptation}
\label{app:evaluation}

\nbf{Browser evaluation.}
For WebRise~\citep{meng2026webrise}, all generated pages are evaluated with the same Interaction Contract Graph and browser executor used to construct the execution rewards in~\cref{sec:loop_formulation}.
The evaluator executes the prescribed interactions and assertions and returns transition validity, explicit and implicit requirement coverage, and requirement-level execution labels.
For WebGen-Bench~\citep{lu2025webgenbench}, we use the benchmark-provided evaluation harness and disable automatic starter-template selection (\texttt{autoSelectTemplate=false}), requiring each model to construct the website from an empty workspace.
Pages that fail to launch are reported as Start-Fail separately from functional failures after initialization.

\nbf{Same-backbone controls.}
The \emph{refine (w/o RL)} baseline uses the original Qwen3.5 checkpoint for Generator, Critic, and Refiner and performs one generate--critique--refine cycle without parameter updates.
Generation Only optimizes only the initial generation branch.
Refine without Critique trains generation and a second-pass refinement without learned critique guidance.
Staged Training first optimizes generation and subsequently trains critique and refinement, allowing direct comparison with WebLoop's joint optimization.

\nbf{Critique-learning baselines.}
We additionally adapt Critique-GRPO~\citep{critique_grpo} and Critique-RL~\citep{Critique-RL} to functional Web generation.
For Critique-GRPO, a fixed teacher receives requirement-level execution information during training and provides feedback for policy optimization.
For Critique-RL, the Generator and Refiner actor is kept fixed while only the Critic is optimized, following its actor--critic separation.
At evaluation time, the learned Critic provides feedback that conditions refinement by the same underlying actor.
All controlled variants use the same task distribution and browser evaluator as WebLoop wherever applicable.

\nbf{Evaluation consistency.}
All same-backbone methods are initialized from the same Qwen3.5 checkpoints and evaluated on identical benchmark instances with the same browser infrastructure and metric computation.
For WebLoop, ``gen'' denotes the first-pass output of the trained shared policy, while ``refine'' denotes the output after one critique-conditioned refinement cycle.
External open-source, proprietary, and Web-specialized models use the same benchmark-specific execution harness while retaining the inference interfaces supported by their released implementations.


\section{Executable Training Data Construction Details}
\label{app:data}

We provide additional details on the construction of the executable Web training corpus introduced in~\cref{sec:data_construction}.
The objective is to pair natural-language task specifications with browser-executable supervision that captures both whether a requested function is present and whether its behavior remains correct across interactions.
This construction provides page-level execution signals for the Generator and Refiner and requirement-level supervision for Critic learning.

\subsection{Task and Requirement Construction}
\label{app:data_requirement}

\nbf{Domain and scenario coverage.}
We construct the corpus around functional Web applications rather than isolated visual layouts.
The task space spans 21 domains, including e-commerce, office productivity, education, developer tools, and data analytics.
Within each domain, we instantiate concrete scenarios covering core operations, management functions, user settings, data presentation, and multi-step workflows.
The resulting corpus contains 1,442 scenarios, which are further instantiated into 2,880 executable Web generation tasks.

\nbf{Functional task instantiation.}
Each task specifies a target Web application through a natural-language description together with a set of functional requirements.
Following the notation in the main text, we write a task as
\begin{equation}
x=(d,\mathcal{R}_x),
\end{equation}
where $d$ denotes the task description and $\mathcal{R}_x$ contains the requirements that define the expected behavior.
Unlike page-generation data centered primarily on appearance or static content, these requirements describe interactive functionality that can subsequently be tested through browser execution.
This makes each task suitable for evaluating both the initial implementation and its later refinement under the same functional specification.

\nbf{Explicit and implicit requirements.}
We decompose the requirement set into explicit and implicit components,
\begin{equation}
\mathcal{R}_x
=
\mathcal{R}_x^{\mathrm{exp}}
\cup
\mathcal{R}_x^{\mathrm{imp}}.
\end{equation}
Explicit requirements describe functions and interaction entry points directly requested by the task specification.
They capture what operations the generated page should expose and provide the primary requirement-level targets inspected by the Critic.
Implicit requirements instead describe behavioral constraints needed for these functions to operate consistently, including state consistency, operation ordering, boundary handling, and reset behavior.
They therefore extend supervision beyond the presence of an interface element to the correctness of the behavior induced by interacting with it.

The final corpus contains an average of 11.87 requirements per task, including 5.47 explicit and 6.39 implicit requirements.
Implicit requirements account for 53.9\% of all requirements, reflecting the importance of interaction semantics that are not fully characterized by explicit function descriptions alone.
This decomposition also supports the separate explicit- and implicit-requirement coverage terms used by the execution evaluator.

\subsection{Interaction Contract Construction and Executability Filtering}
\label{app:data_icg}

\nbf{From requirements to executable contracts.}
Following the executable evaluation formulation of WebRISE~\citep{meng2026webrise}, we convert each task specification into an initial-state contract and an Interaction Contract Graph (ICG).
The initial-state contract defines the starting condition under which the subsequent browser interactions are executed.
The ICG then operationalizes the functional requirements as executable interaction trajectories and corresponding assertions.
In this way, natural-language requirements are transformed into programmatically verifiable behaviors rather than evaluated only through textual or visual similarity.

\nbf{Interaction Contract Graph.}
For a task $x$, its ICG $G_x$ organizes browser execution through states, transitions, and functional assertions.
Transitions describe the interactions used to move between Web states, while assertions verify whether the resulting behavior satisfies the corresponding functional constraints.
A single task can therefore contain multiple interaction transitions and multiple assertions, enabling evaluation of multi-step functionality rather than only isolated final states.
Across the corpus, each ICG contains 10.64 states, 11.40 transitions, and 28.15 executable assertions on average.

Executing a generated implementation $y$ against $G_x$ produces the execution signals used throughout WebLoop,
\begin{equation}
E(y;G_x)
=
\left(
T(y),
R_e(y),
R_i(y),
\mathbf{z}(y)
\right),
\end{equation}
where $T$ measures transition validity, $R_e$ and $R_i$ measure explicit- and implicit-requirement coverage, and $\mathbf{z}$ provides requirement-level execution labels.
The first three quantities form the page-level reward used for Generator and Refiner optimization, while the requirement-level labels provide execution-grounded supervision for Critic discriminability.
Thus, the same executable contract supports both outcome-level optimization and intermediate Critic credit assignment.

\nbf{Reference implementation and browser validation.}
To verify executability and reduce dependence on any single generator, we independently construct reference HTML implementations using Kimi-K2.6, Claude Opus 5, and GPT-5, and execute each against the associated ICG in a real browser.
We retain a task only if all three implementations achieve transition validity of at least $0.60$:
\begin{equation}
\small
T_m \geq 0.60,
\qquad
\forall m\in\{1,2,3\},
\end{equation}
where $T_m$ denotes the transition validity of the implementation generated by the $m$-th model.
This cross-model criterion filters tasks with insufficiently reliable interaction contracts before training.

\nbf{Human review and filtering.}
Browser-executable contracts may still be semantically misaligned with the intended requirements, for example by omitting necessary prerequisites or using assertions that execute successfully but do not faithfully test the target behavior.
We therefore conduct a manual audit of sampled candidate tasks with two reviewers.
For each audited task, the reviewers examine five aspects:
(1) \emph{requirement alignment}, whether each tested behavior is supported by an explicit requirement or a justified implicit constraint;
(2) \emph{interaction coherence}, whether the action sequence includes the necessary preconditions and follows a valid progression toward the target state;
(3) \emph{assertion validity}, whether the expected outcomes correctly reflect the intended functional semantics;
(4) \emph{structural consistency}, whether states, transitions, and requirement references are internally consistent with the ICG specification; and
(5) \emph{functional discriminability}, whether the checks distinguish actual functional completion from superficial interface presence.
Audited tasks containing invalid or misaligned interaction contracts are excluded before corpus finalization.

\nbf{Final corpus.}
After executable validation and manual filtering,, the training corpus contains 2,880 tasks across 21 domains and 1,442 scenarios.
Together, their explicit and implicit requirements and associated ICGs provide fine-grained supervision over both functional outcomes and intermediate interaction behavior.
This construction turns each natural-language Web task into an executable training unit that can evaluate initial generation, diagnose requirement-level failures, and measure the downstream effect of critique-conditioned refinement.

\begin{table}[t]
\centering
\small
\setlength{\tabcolsep}{4pt}
\renewcommand{\arraystretch}{1.12}
\caption{
\textbf{Human validation of functional evaluation.}
Agreement between the automatic evaluator and two independent human reviewers on the same 300 sampled interaction cases.
Human--human agreement is included as a reference.
}

\label{tab:human_consistency}

\begin{tabularx}{\linewidth}{@{}Xrrrrrr@{}}
\toprule
\textbf{Comparison}
& \textbf{\#Samples}
& \textbf{Acc.}$\uparrow$
& \textbf{MAE}$\downarrow$
& \textbf{Spearman}$\uparrow$
& \textbf{Pearson}$\uparrow$
& $\boldsymbol{\kappa}\uparrow$ \\
\midrule
Human (A) vs.\ Human (B)
& 300 & 0.920 & 0.079 & 0.924 & 0.923 & 0.838 \\
\midrule
Evaluator vs.\ Human (A)
& 300 & 0.880 & 0.095 & 0.901 & 0.902 & 0.758 \\
Evaluator vs.\ Human (B)
& 300 & 0.900 & 0.094 & 0.902 & 0.898 & 0.799 \\
\bottomrule
\end{tabularx}
\end{table}

\subsection{Human Consistency Validation}
\label{app:data_human_validation}

\nbf{Annotation setup.}

We further assess the reliability of the automatic functional evaluator through human consistency analysis on 300 sampled interaction cases.
For each case, two reviewers independently judge whether the expected behavior is satisfied based on the task requirements, tested interaction, and corresponding execution evidence.

\nbf{Metrics.}

We report accuracy, mean absolute error (MAE), Spearman correlation,
Pearson correlation, and Cohen's $\kappa$.
Accuracy and Cohen's $\kappa$ measure agreement on binary pass/fail
judgments.
MAE and the correlation coefficients are computed over paired
human and automatic graded scores normalized to $[0,1]$.
As shown in~\cref{tab:human_consistency}, the evaluator shows strong consistency with both reviewers across accuracy, MAE, rank and linear correlation, and Cohen's $\kappa$, while human--human agreement provides an upper-reference level.
These results indicate that the automatic evaluator closely tracks human judgments and provides a reliable execution-based supervision signal for WebLoop training and evaluation.

\section{Additional Experimental Results and Analysis}
\label{app:additional_results}

\begin{figure}[!t]
\centering
\includegraphics[width=\linewidth]{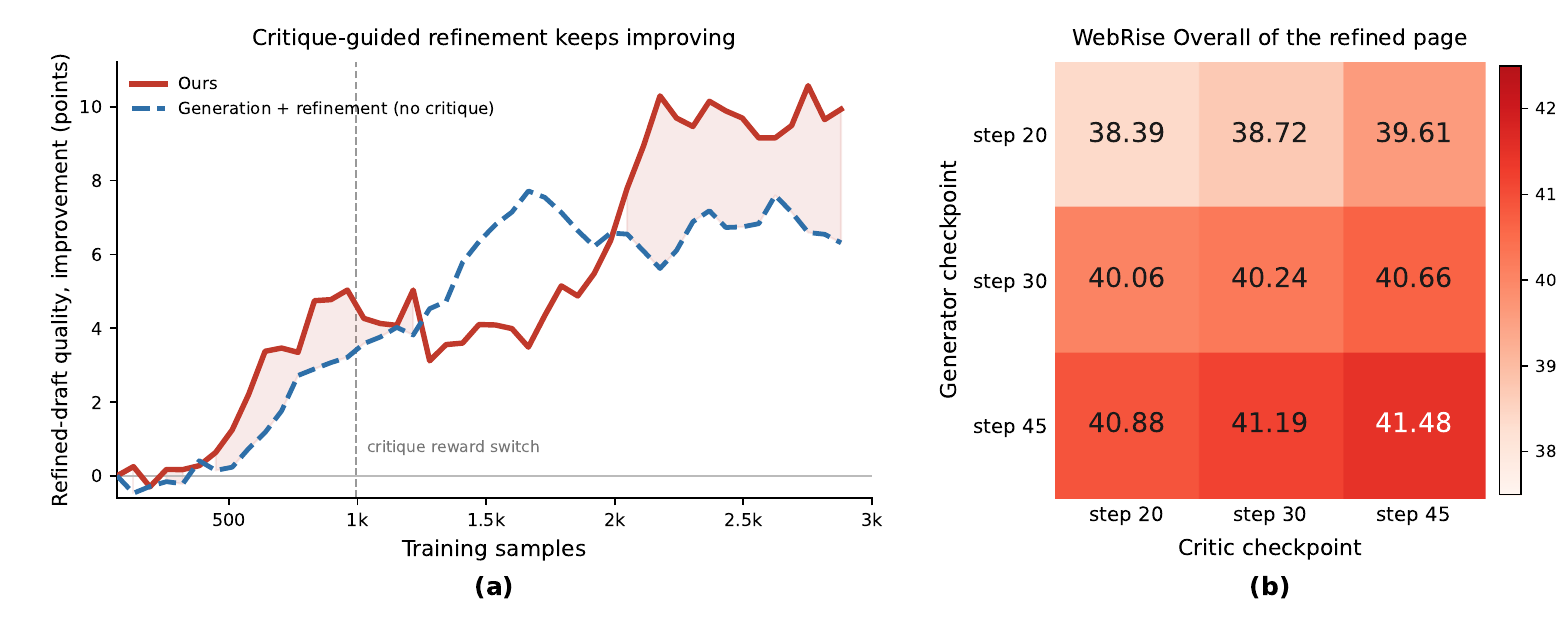}
\caption{
\textbf{Refinement learning and Generator--Critic checkpoint interaction.}
(a) Refined-page quality over training, reported as improvement from the initial training point, with refinement without learned critique as a control.
The dashed line marks the Critic reward switch.
(b) WebRise Overall under cross-pairing Generator and Critic checkpoints from steps 20, 30, and 45. The Refiner uses the same checkpoint as the Generator in each pairing.
}
\label{fig:loop_cadaptation}
\end{figure}

\subsection{Training Dynamics and Loop Co-Adaptation}
\label{app:training_dynamics}

\nbf{Refinement gains over training.}
\Cref{fig:loop_cadaptation}(a) compares WebLoop with generation and refinement trained without a learned Critic.
Both improve with training, but critique-guided refinement develops a larger advantage, particularly after downstream helpfulness is introduced.
This trend suggests that WebLoop increasingly exploits learned intermediate feedback rather than benefiting only from an additional generation pass.

\nbf{Generator--Critic checkpoint interaction.}
\Cref{fig:loop_cadaptation}(b) cross-pairs Generator and Critic checkpoints to separate their contributions over training.
For every fixed Generator, later Critic checkpoints improve refinement quality; likewise, later Generators improve performance under every fixed Critic.
The score rises from 38.4 for the step-20 pair to 41.5 for the step-45 pair, with the best result obtained when both use the latest checkpoint.
These results indicate that both roles improve progressively and contribute jointly to the final refinement gain.

\subsection{Fine-Grained Functional Evaluation}
\label{app:fine_grained_eval}

\nbf{WebGen-Bench category breakdown.}
\Cref{tab:webgen_categories} decomposes WebGen-Bench along two independent taxonomies provided by the benchmark.
Website categories group checks into Content Presentation, User Interaction, and Data Management, while check categories distinguish Functional Testing, Data Display Testing, and Design Validation.
Each taxonomy covers all 647 checks, so the two groups provide complementary views of performance rather than six mutually exclusive categories.
We use the same template-free evaluation protocol and scoring rule as in the main results.

\begin{table}[!t]
\caption{
\textbf{Fine-grained WebGen-Bench results.}
Accuracy under the same template-free setting as \cref{tab:mainresult}.
The 647 interaction checks are independently grouped by website category and check category; each grouping covers the full benchmark, so the six columns are not additive.
$^\dagger$ denotes proprietary models.
}
\label{tab:webgen_categories}
\begin{center}
\small
\setlength{\tabcolsep}{4pt}
\renewcommand{\arraystretch}{1.08}
\resizebox{\textwidth}{!}{
\begin{tabular}{l|ccc|ccc}
\toprule
\multirow{2}{*}{\textbf{Model / Method}}
& \multicolumn{3}{c|}{\textbf{By Website Category}}
& \multicolumn{3}{c}{\textbf{By Check Category}} \\
\cmidrule(lr){2-4}
\cmidrule(lr){5-7}
& Content & User & Data
& Functional & Data Display & Design \\
& Presentation & Interaction & Management
& Testing & Testing & Validation \\
\midrule

\multicolumn{7}{l}{\textit{General-Purpose Models}} \\
\midrule

Qwen3.5-122B-A10B
& 21.3 & 9.6 & 10.6
& 5.2 & 18.0 & 27.0 \\

Kimi-K2.6
& 42.5 & 26.4 & 28.4
& 20.9 & 37.9 & 49.6 \\

Qwen3.5-397B-A17B
& 27.3 & 19.2 & 16.2
& 11.5 & 24.5 & 40.2 \\

GLM-5.3
& 26.1 & 25.9 & 28.1
& 20.2 & 25.8 & 45.1 \\

Claude Opus 4.7$^\dagger$
& \textbf{67.0} & \textbf{45.8} & \textbf{61.3}
& \textbf{41.3} & \textbf{69.1} & \textbf{73.4} \\

Gemini 3.1 Pro$^\dagger$
& 46.6 & 45.4 & 48.4
& 35.4 & 50.5 & 70.9 \\

GPT-6 Astra$^\dagger$
& 48.6 & 39.0 & 45.6
& 30.5 & 51.1 & 66.4 \\

\midrule
\multicolumn{7}{l}{\textit{Web-Generation Specialized Models}} \\
\midrule

WebGen-LM-7B
& 2.0 & \textbf{5.4} & 2.5
& \textbf{1.6} & \textbf{4.8} & 8.2 \\

WebGen-Agent-7B (Step-GRPO)
& \textbf{6.0} & 5.0 & \textbf{5.6}
& 0.4 & 3.8 & \textbf{21.7} \\

WebGen-LM-32B
& 2.9 & 4.3 & 1.9
& 1.2 & 3.8 & 8.6 \\

UIGEN-T2-7B
& N/A & N/A & N/A
& N/A & N/A & N/A \\

WebGen-LM-14B
& 1.7 & 5.0 & 2.5
& 1.3 & \textbf{4.8} & 7.4 \\

\midrule
\multicolumn{7}{l}{\textit{Controlled Comparisons: Qwen3.5}} \\
\midrule

Base (Qwen3.5-9B)
& 29.0 & 17.1 & 30.0
& 15.5 & 25.0 & 43.4 \\

\quad Generate--Critique--Refine (w/o RL)
& 42.0 & 24.0 & 33.1
& 22.1 & 34.7 & 50.4 \\

\quad Critique-GRPO
& 12.6 & 17.1 & 24.1
& 11.1 & 16.9 & 36.9 \\

\quad Critique-RL
& 28.7 & 28.3 & 28.7
& 19.5 & 30.4 & 50.8 \\

\quad \textbf{WebLoop} (Gen)
& 41.1 & \textbf{31.2} & 32.5
& 23.0 & 37.9 & \textbf{59.4} \\

\quad \textbf{WebLoop} (Refine)
& \textbf{50.3} & 30.2 & \textbf{43.4}
& \textbf{27.3} & \textbf{47.0} & 58.6 \\

\cmidrule(lr){1-7}

Base (Qwen3.5-27B)
& 45.1 & 23.6 & 25.0
& 17.1 & 39.5 & 50.0 \\

\quad Generate--Critique--Refine (w/o RL)
& 48.6 & 33.2 & \textbf{36.2}
& 24.0 & 50.3 & 58.6 \\

\quad \textbf{WebLoop} (Gen)
& 43.4 & 27.6 & 30.0
& 19.0 & 41.7 & 55.7 \\

\quad \textbf{WebLoop} (Refine)
& \textbf{59.2} & \textbf{37.1} & \textbf{36.2}
& \textbf{29.5} & \textbf{51.3} & \textbf{66.8} \\

\bottomrule
\end{tabular}
}
\end{center}
\end{table}

\nbf{Where does refinement help?}
As shown in~\cref{tab:webgen_categories}, WebLoop improves functional behavior across a broad range of categories rather than only interface-level compliance.
For Qwen3.5-9B, the final Refine output improves over the base model by 11.8 points on Functional Testing and 22.0 points on Data Display Testing, while also increasing all three website-category scores.
The 27B model shows the same pattern, including gains of 12.4 points on Functional Testing and 11.8 points on Data Display Testing.
Compared with first-pass WebLoop generation, refinement is more heterogeneous at 9B: it substantially improves Content Presentation, Data Management, Functional Testing, and Data Display Testing, while User Interaction and Design Validation decrease slightly by 1.0 and 0.8 points.
At 27B, refinement improves all six categories, suggesting that larger models exploit critique-conditioned revision more consistently.

\begin{figure}[!t]
\centering
\includegraphics[width=\textwidth]{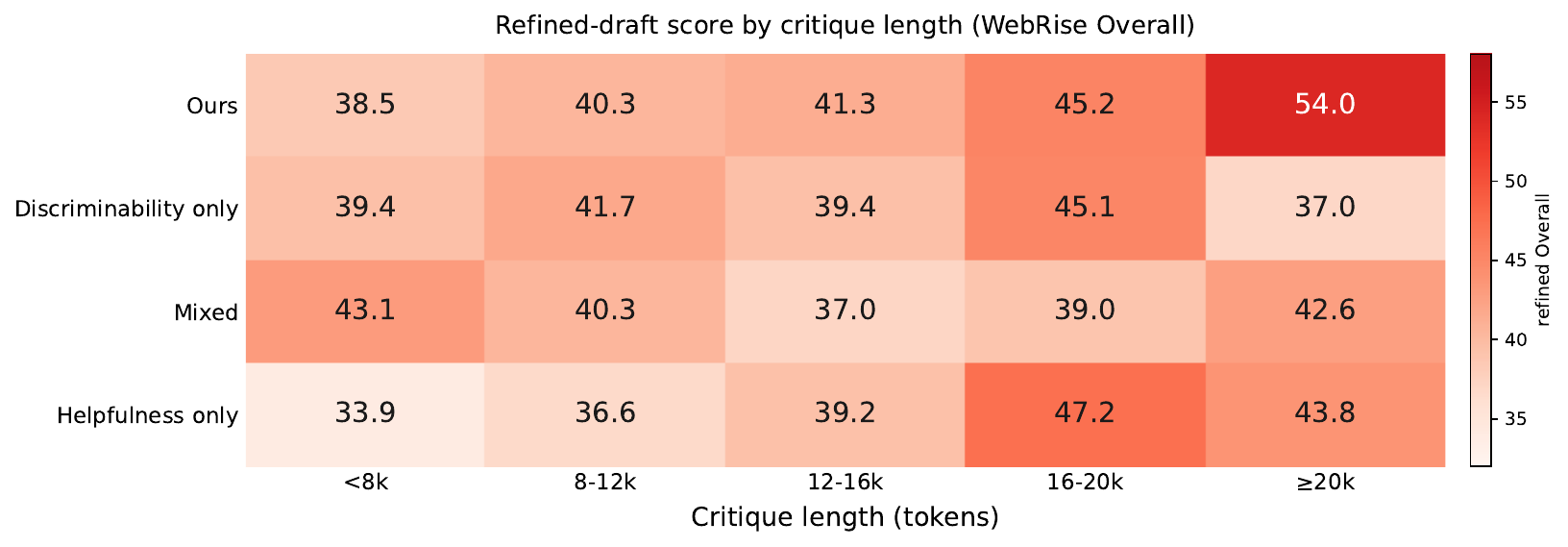}
\caption{
\textbf{Critique length versus refinement quality.}
Each cell reports the mean WebRise Overall of refined pages within a critique-length bucket for a given Critic objective.
WebLoop exhibits a stronger positive association, while alternative objectives show weaker or non-monotonic trends.
}
\label{fig:critlen}
\end{figure}

\subsection{Critique Diagnostics and Repair Analysis}
\label{app:critique_analysis}

\nbf{Critique length and refinement quality.}
\Cref{fig:critlen} examines the association between critique length and downstream refinement quality under different Critic objectives.
For WebLoop, refined WebRise Overall increases from 38.5 in the shortest bucket to 54.0 in the longest, whereas the alternative objectives show weaker or non-monotonic patterns.
This contrast suggests that critique utility depends on how additional feedback is structured and used, rather than on verbosity alone.
Because critique length is neither controlled nor independently randomized, we treat this analysis as descriptive rather than causal.

\begin{figure}[!tp]
\centering
\includegraphics[width=\linewidth]{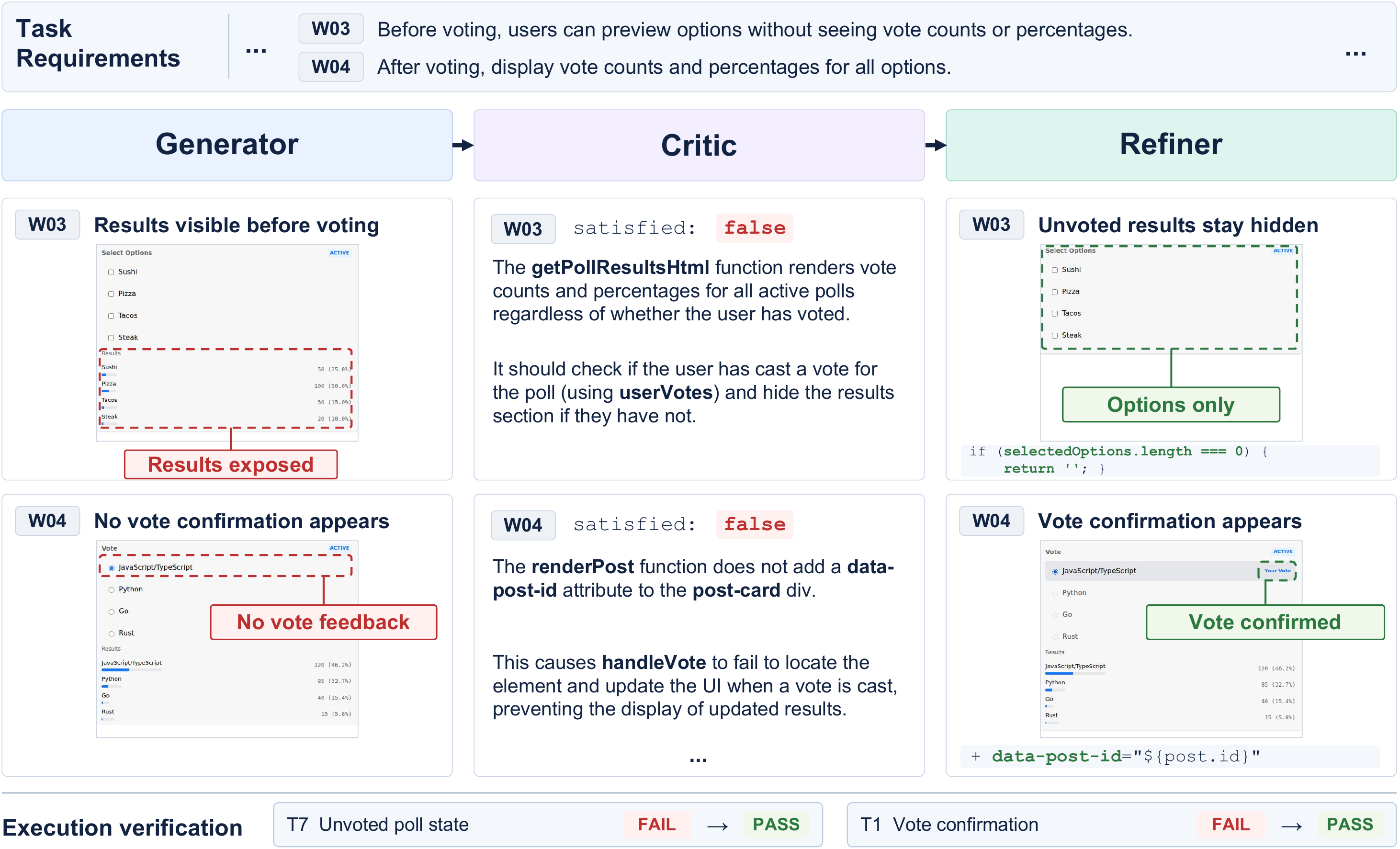}
\caption{
\textbf{Successful critique-guided repair on Poll Vote (WebRise).}
The Critic localizes premature disclosure of voting results and a missing post identifier that blocks vote updates.
The Refiner applies the corresponding fixes, changing both affected transition checks from FAIL to PASS.
}
\label{fig:case_study}
\end{figure}

\subsection{Qualitative Analysis of the Improvement Loop}
\label{app:qualitative_analysis}

We utilize qualitative examples to examine where the improvement loop succeeds or breaks down between requirement diagnosis and executable repair.
We organize the analysis around successful repair and three distinct sources of failure: incorrect Critic judgments, insufficient repair guidance, and failures in Refiner implementation.

\nbf{Successful diagnosis and repair.}
\Cref{fig:case_study} illustrates how an execution-free Critic can support targeted functional repair.
For W03, the Critic identifies that voting results are exposed before a vote is cast and localizes the missing state-dependent guard.
For W04, it traces the failed post-vote update to a missing \texttt{data-post-id} attribute.
Conditioned on these diagnoses, the Refiner applies targeted code changes that turn both associated browser checks from FAIL to PASS.
This example demonstrates how requirement-level diagnosis can yield actionable guidance and executable improvement.

\begin{figure}[!tp]
\centering
\includegraphics[width=\linewidth]{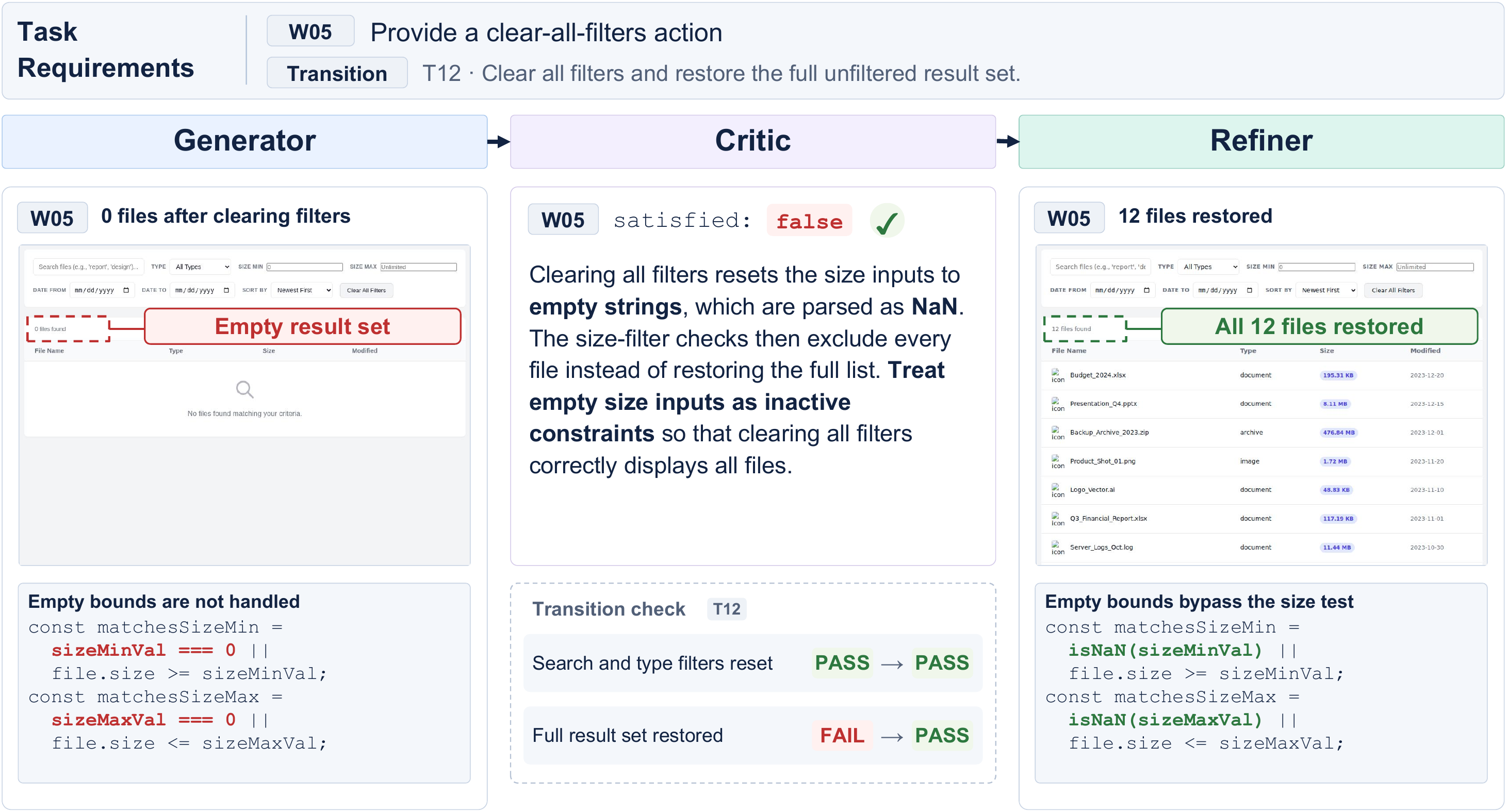}
\caption{
\textbf{Boundary-condition repair on File Search with Filters (WebRise).}
The Critic traces the empty result set to size bounds parsed as \texttt{NaN} after clearing filters.
The Refiner adds guards for empty bounds, restoring all 12 files while preserving the already-correct control reset.
}
\label{fig:bench_success}
\end{figure}

The file-filter example in~\cref{fig:bench_success} provides a complementary boundary-condition case.
Although clearing the controls already works, empty size inputs are parsed as \texttt{NaN}, causing all files to be excluded.
The Critic localizes this defect, and the Refiner adds guards that restore all 12 files while preserving the correct control-reset behavior.
Across both examples, the diagnosed defect, implemented change, and browser-verified improvement are well aligned.

\begin{figure}[!tp]
\centering
\includegraphics[width=\linewidth]{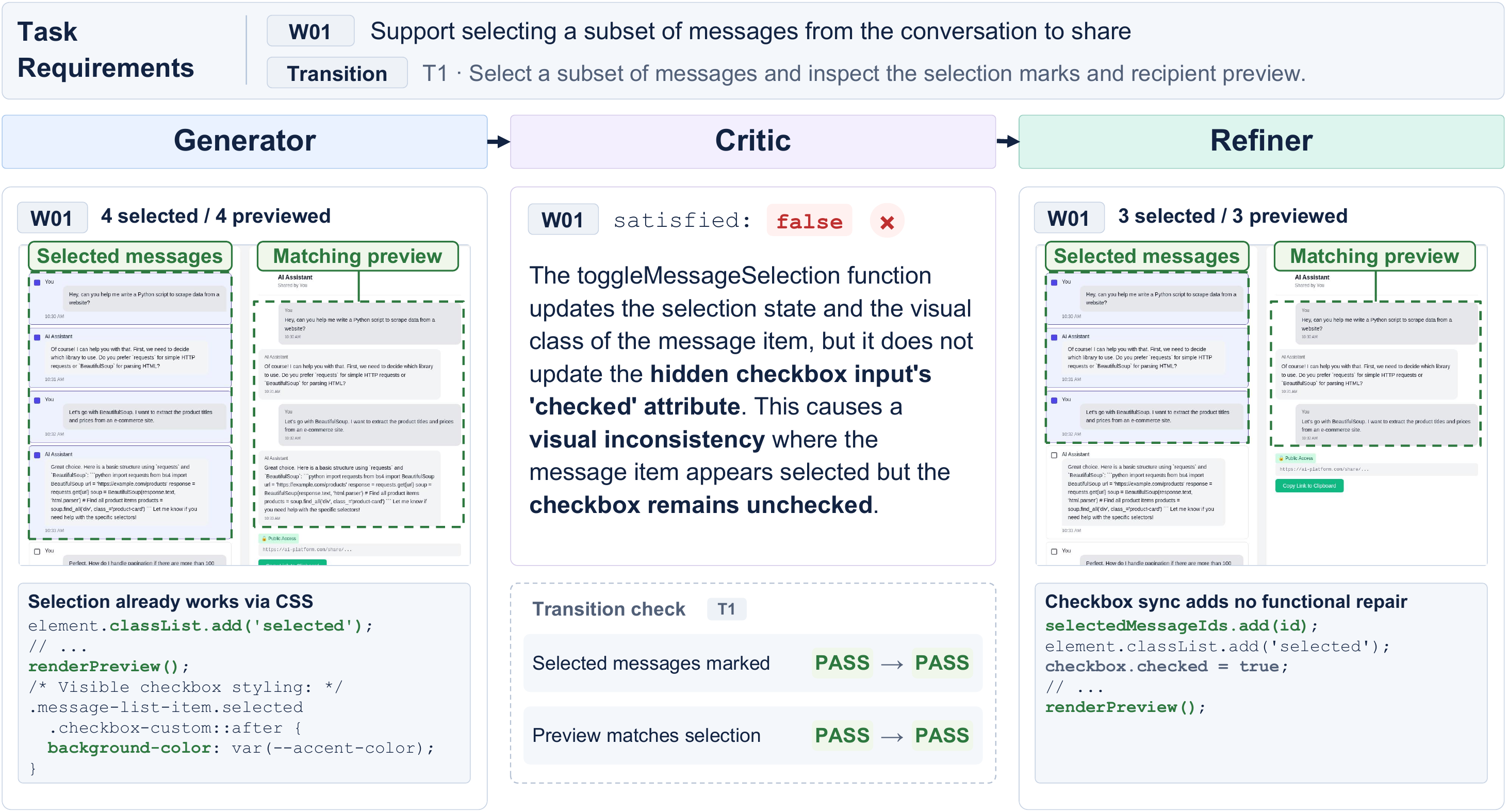}
\caption{
\textbf{False-positive Critic judgment on Share Chat Link Interface (WebRise).}
The Critic attributes message-selection failure to an unsynchronized hidden checkbox, although the visible selection and preview are already controlled correctly.
The Refiner synchronizes the checkbox without affecting either passing behavior.
}
\label{fig:bench_false_alarm}
\end{figure}

\nbf{Critic judgment errors.}
The Critic may incorrectly classify already functional behavior as violating a requirement.
In~\cref{fig:bench_false_alarm}, it attributes broken message selection to a stale hidden-checkbox property.
However, a CSS class controls the visible selection marker and a message-ID set controls the preview, and both behaviors already pass in the draft.
The Refiner synchronizes the hidden checkbox while preserving these mechanisms, so the relevant predicates remain PASS.
This case therefore reflects a false-positive Critic judgment rather than a failed Refiner implementation: preserving correct behavior after revision does not validate the preceding negative verdict.

\begin{figure}[!tp]
\centering
\includegraphics[width=\linewidth]{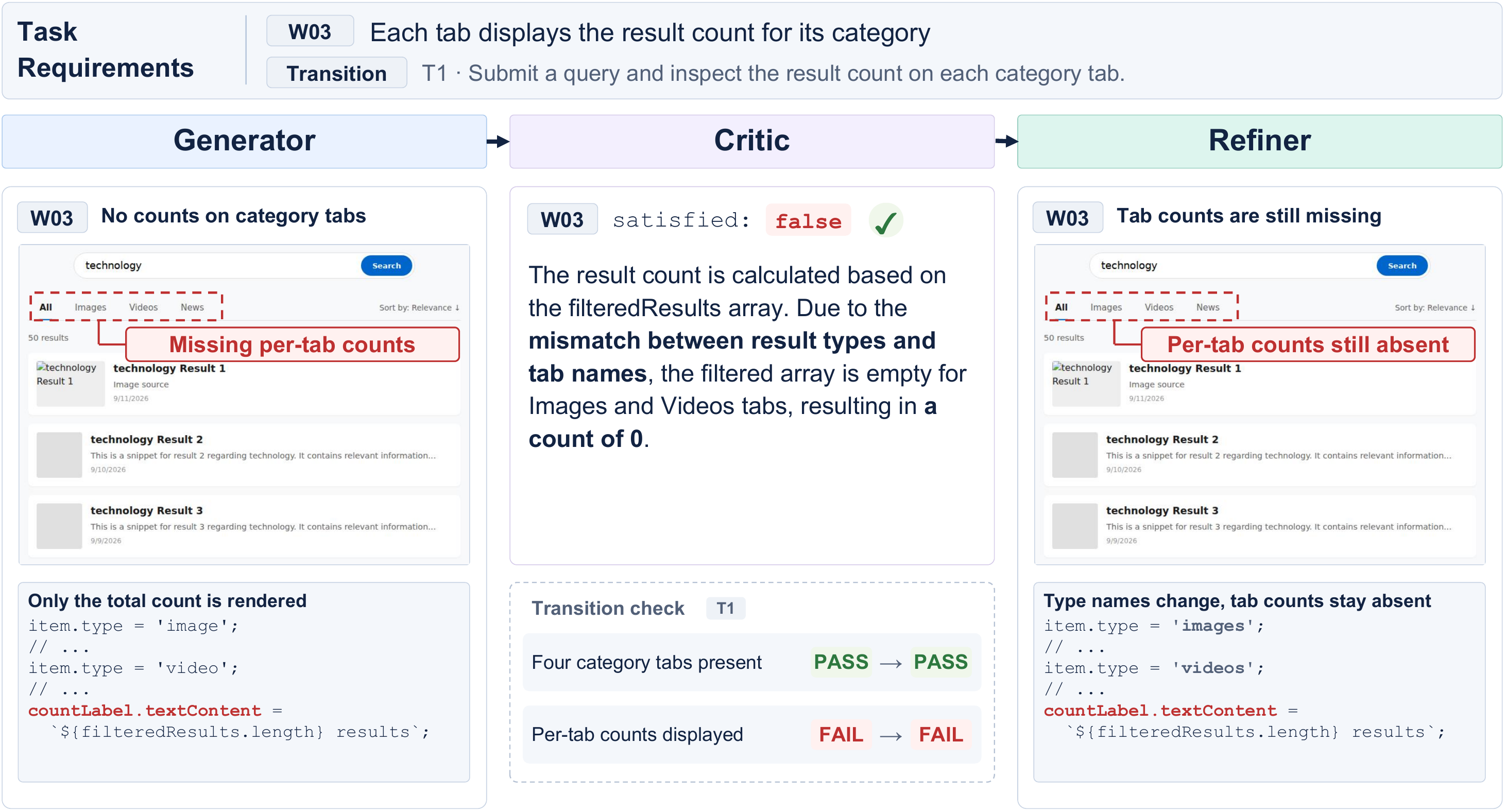}
\caption{
\textbf{Correct verdict but mislocalized repair guidance on Search Result Tabs (WebRise).}
The Critic correctly marks the per-category count requirement as unmet but targets a type-matching issue rather than the missing count labels.
The Refiner applies the proposed edit, while the target predicate remains FAIL.
}
\label{fig:bench_misdirected}
\end{figure}

\nbf{Limitations of Critic repair guidance.}
Even when the requirement verdict is correct, the proposed repair may fail to address the defect responsible for the unmet behavior.
In~\cref{fig:bench_misdirected}, the Critic identifies a genuine type-matching issue but does not address the required per-tab counts.
The Refiner applies the suggested edit, yet the counts remain absent.
The failure therefore reflects insufficient repair guidance rather than failure to follow the recommendation.

\begin{figure}[!tp]
\centering
\includegraphics[width=\linewidth]{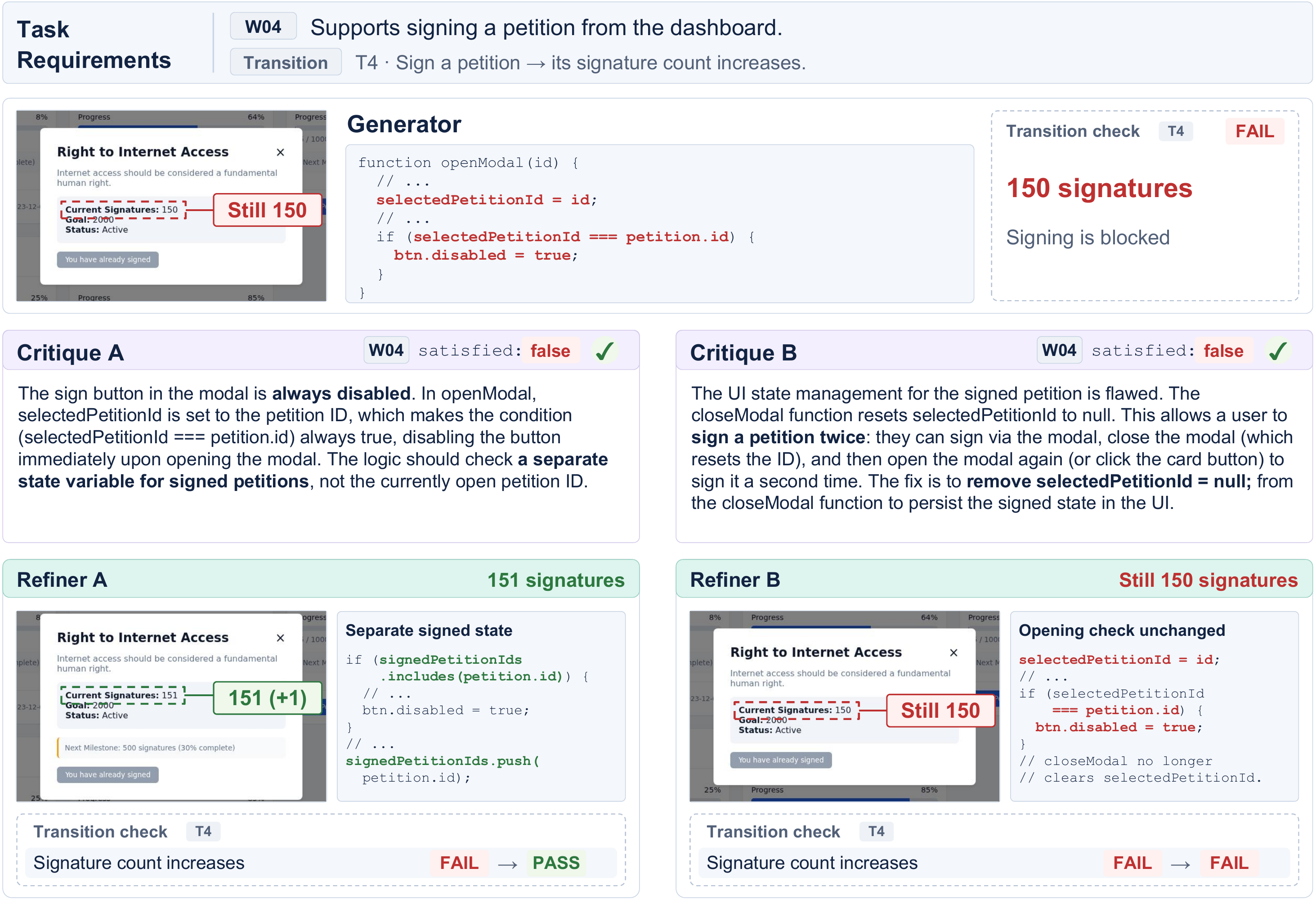}
\caption{
\textbf{Divergent fault localization on Petition Progress Dashboard (WebRise).}
Both critiques correctly identify signing as broken, but only one localizes the state error that disables signing when the dialog opens.
Although both suggested edits are implemented, only the correctly localized repair changes the transition from FAIL to PASS.
}
\label{fig:pair_fault_localization}
\end{figure}

The paired critiques in~\cref{fig:pair_fault_localization} further isolate this distinction.
They share the same draft, requirement verdicts, and discriminability score ($D=0.75$), but differ in fault localization.
Only Critique A identifies the conflation of petition-opening and signing states.
Its corresponding refinement restores the disabled signing behavior, whereas Critique B's modal-close edit is implemented without removing the blocker.
Thus, verdict-level correctness alone is insufficient to establish the correctness of the proposed repair direction.

\begin{figure}[!tp]
\centering
\includegraphics[width=\linewidth]{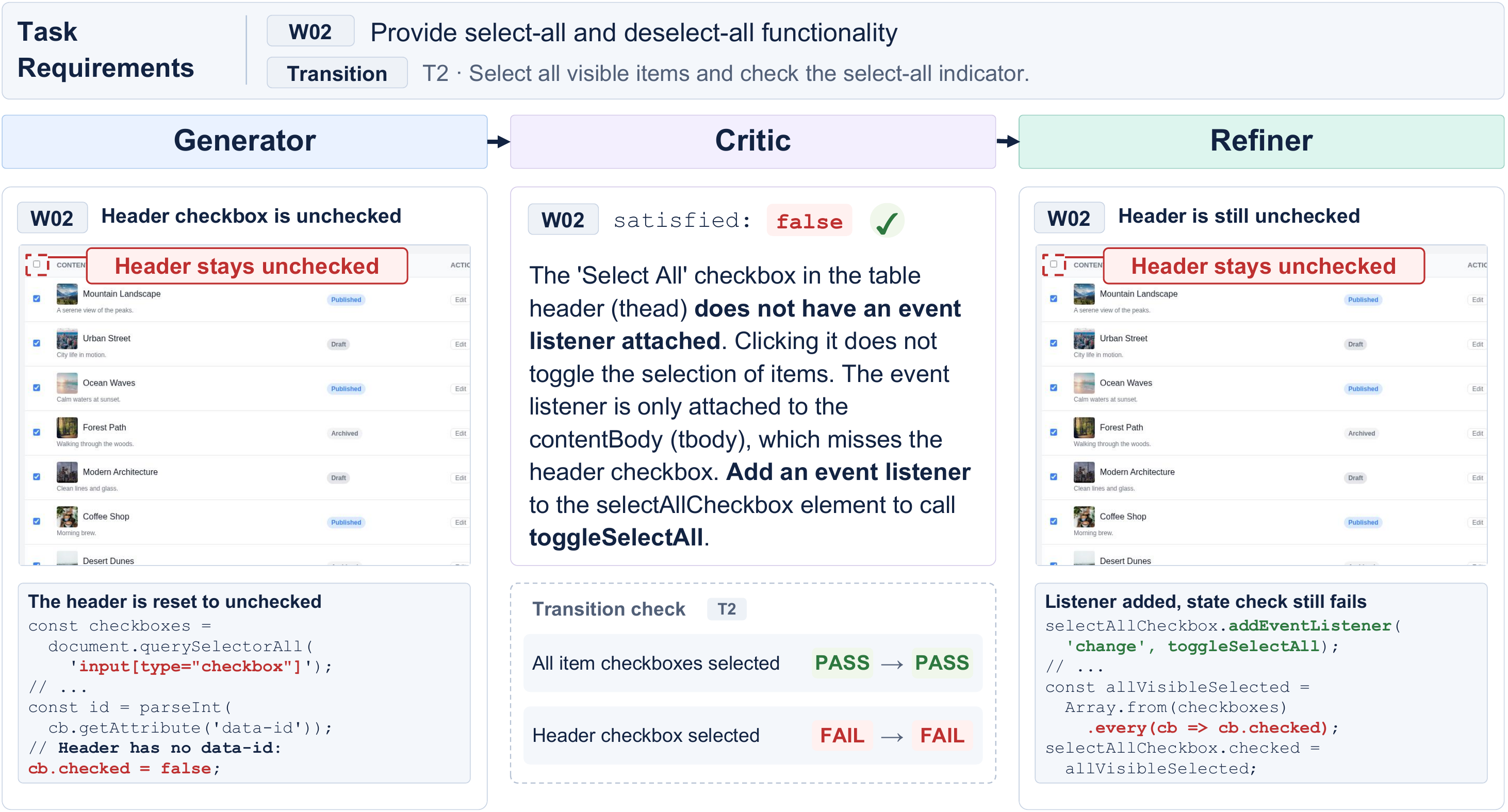}
\caption{
\textbf{Incomplete repair guidance on Bulk Content Operations (WebRise).}
The Refiner adds the missing listener as recommended, but an unaddressed selector error still clears the header-checkbox state.
The prescribed edit is implemented, yet the target requirement remains unsatisfied.
}
\label{fig:bench_incomplete}
\end{figure}

Repair guidance may also be valid but incomplete.
In~\cref{fig:bench_incomplete}, the Critic identifies a missing listener and the Refiner adds it, but an unaddressed selector error leaves the header checkbox unchecked.
The prescribed edit is implemented correctly, yet the target behavior remains unsatisfied because the guidance does not cover all contributing defects.

\begin{figure}[!tp]
\centering
\includegraphics[width=\linewidth]{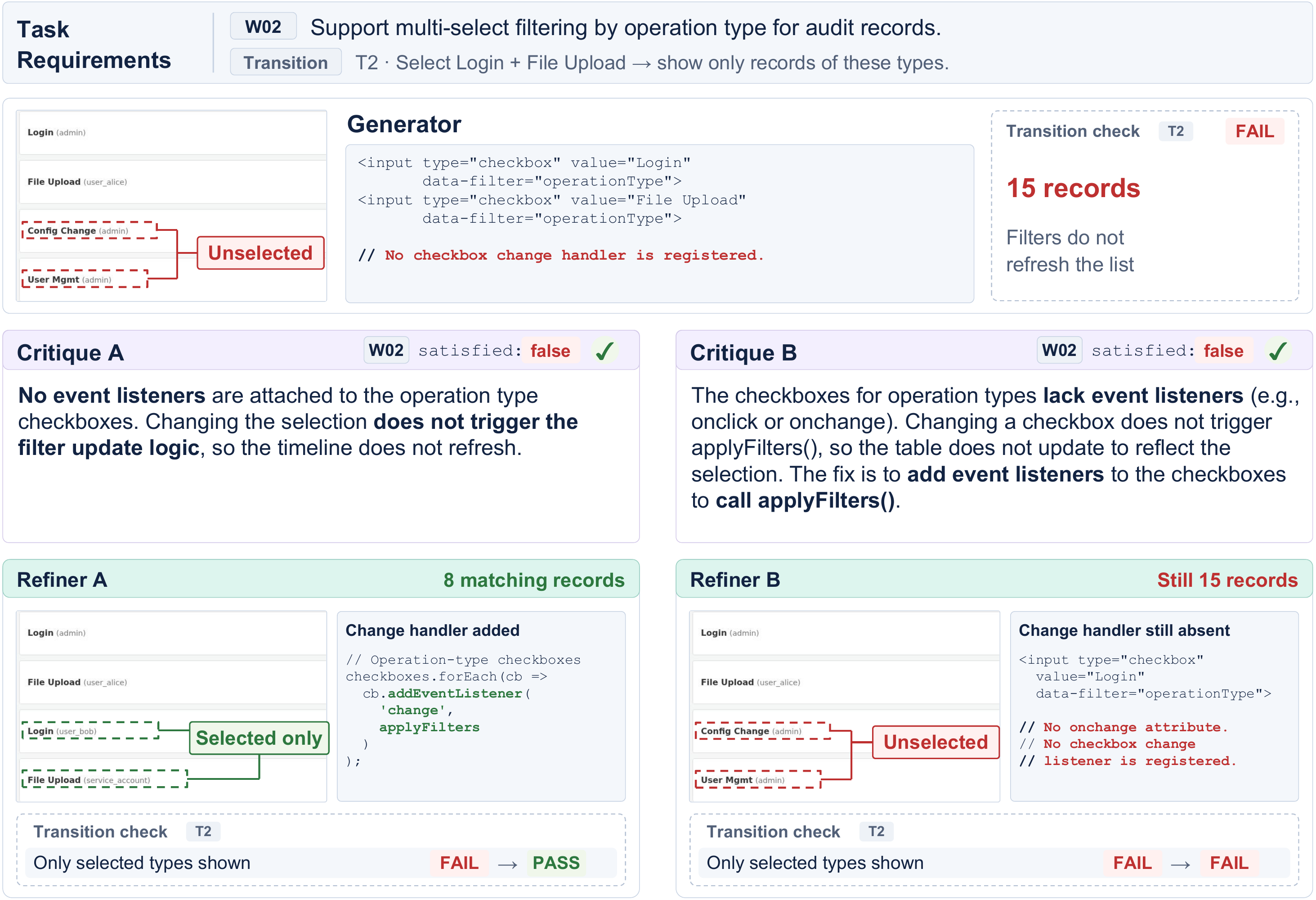}
\caption{
\textbf{Divergent Refiner outcomes under equivalent diagnoses on Audit Timeline (WebRise).}
Both critiques identify the same missing event binding, but only one refinement implements the recommended listener and passes the transition check.
The contrast isolates implementation fidelity from Critic diagnosis.
}
\label{fig:pair_implementation}
\end{figure}

\nbf{Refiner implementation failures.}
Failure can instead arise when the Refiner does not faithfully realize an appropriate recommendation.
In~\cref{fig:pair_implementation}, both critiques correctly identify the missing event binding and explicitly recommend adding it, but one displayed refinement leaves the binding absent and fails.
Another refinement conditioned on the same critique adds the binding and passes.
Unlike the incomplete-guidance cases above, the required repair is already specified here; the failure lies in executing that guidance.

\begin{figure}[!tp]
\centering
\includegraphics[width=\linewidth]{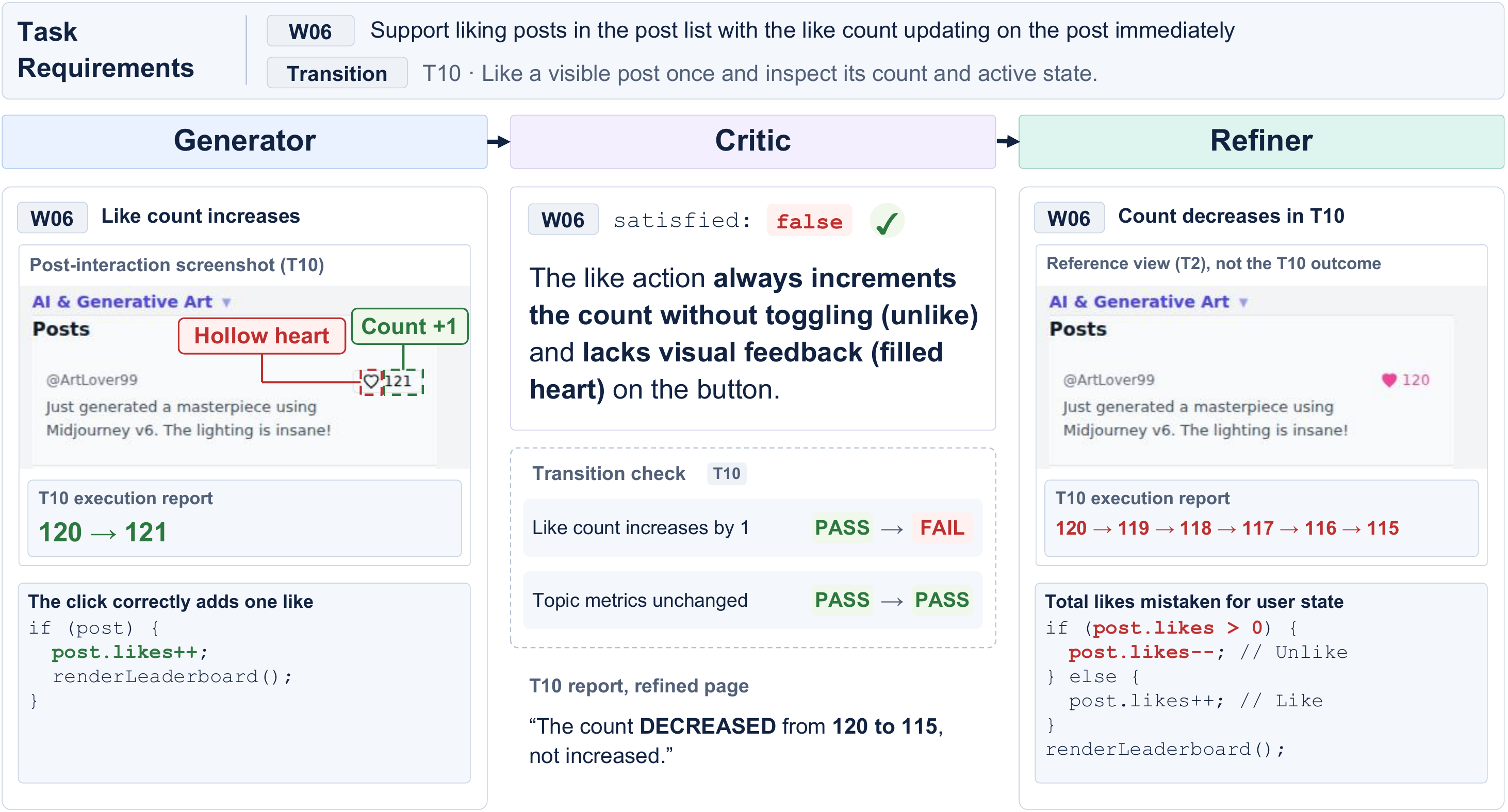}
\caption{
\textbf{Refiner-induced regression on Topic Trending (WebRise).}
While addressing missing liked-state feedback, the Refiner conflates the user's like state with the aggregate like count, causing the count-update predicate to regress from PASS to FAIL.
}
\label{fig:bench_regression}
\end{figure}

The Refiner may also introduce regressions while addressing genuine defects.
In~\cref{fig:bench_regression}, the Critic correctly identifies missing liked-state feedback, but the revision conflates the user's like state with the aggregate like count, causing the count predicate to regress from PASS to FAIL.
Thus, unsuccessful refinement may result either from failing to follow valid guidance or from introducing new implementation errors.

Together, these cases distinguish whether the Critic judges the requirement correctly, whether its guidance is sufficient, and whether the Refiner implements that guidance successfully.
Discriminability evaluates the first, whereas helpfulness captures downstream outcomes jointly determined by feedback and implementation.
A failed refinement therefore does not by itself identify the source of failure; the intermediate diagnosis and resulting code changes are required to localize where the improvement loop breaks down.

\section{Prompt Templates}
\label{app:prompts}

WebLoop utilizes role-specific prompts to instantiate the Generator, Critic, and Refiner within the shared policy.

\nbf{Generator.}
The Generator prompt in~\cref{fig:prompt-generator} directly maps the task specification to a complete self-contained HTML implementation, restricting the output to executable code without auxiliary explanation.

\nbf{Critic.}
The Critic prompt in~\cref{fig:prompt-critic} adopts an execution-free red-team perspective, requiring requirement-level judgments from the specification and source code alone and structured repair guidance for detected failures.

\nbf{Refiner.}
The Refiner prompt in~\cref{fig:prompt-refiner} places diagnosed defects before the prior implementation, encouraging targeted rewrites while preserving unflagged behavior.

\begin{PromptBox}
{Generator}
{Generator prompt for first-pass Web generation.}
{fig:prompt-generator}

You are an expert frontend developer.
Given a task description and a set of functional requirements, produce a single, complete, self-contained HTML file that fulfills all requirements.

\textbf{Rules:}
\begin{enumerate}[nosep, leftmargin=*, itemsep=2pt]
\item Output only the HTML code, starting with \texttt{\textless!DOCTYPE html\textgreater} and ending with \texttt{\textless/html\textgreater}.
\item Do not include any explanation, markdown fences, or commentary.
\item Inline all CSS and JavaScript; no external files.
\item The page must be fully functional on first load, with realistic placeholder data.
\item Implement every explicit requirement completely.
\end{enumerate}

\textbf{Input:} the task statement \texttt{\{input\_text\}}, which already carries the prose requirements and the test data contract, so the generator builds against exactly the statement the evaluation is derived from.

\textbf{Output:} the complete HTML page, with no surrounding prose.
\end{PromptBox}

\begin{PromptBox}
{Critic}
{Critic prompt for requirement-level diagnosis and repair guidance.}
{fig:prompt-critic}

You are a red-team QA engineer. Your explicit goal is to break this self-contained HTML page.
You are given the task it was built for, its explicit requirements (each with an id), and the page.
You have no automated test results --- find breakage by reading the JavaScript.
Assume every requirement is broken until the code proves otherwise; a polished-looking UI is not proof.

\textbf{Where the attacks come from:}
the strongest attacks come from the \emph{implicit constraints} each explicit requirement entails --- the sub-behaviours a careful spec author derives from it:
validation of invalid inputs (negative, zero, empty, non-numeric, out-of-range);
persistence across save, re-selection, and re-render;
consistency while a filter or sort is active or after the data changes;
reversibility (add\,$\leftrightarrow$\,delete, enable\,$\leftrightarrow$\,clear);
and robust DOM handling under repeated or out-of-order use.

\textbf{Procedure, for each explicit requirement:}
\begin{enumerate}[nosep, leftmargin=*, itemsep=2pt]
\item Derive the implicit constraints it entails.
\item Find the input or action sequence most likely to violate one, and confirm the failure in the code.
\item Mark \texttt{satisfied} true only if you could not break the explicit behaviour or any entailed constraint.
\end{enumerate}

\textbf{Output rules:}
\begin{enumerate}[nosep, leftmargin=*, itemsep=2pt]
\item Keep any internal reasoning extremely brief (at most a few sentences per requirement). Do not re-list task data, never re-enumerate every element of an input array, never quote every field, and never repeat a check already made.
\item Do not loop on the same doubt; after one brief re-check, commit to a verdict and move on.
\item Respond with a single JSON object and nothing after it: no reasoning after the JSON, no markdown fences around it.
\item Emit exactly one entry per requirement id given, in the same order, each with a boolean \texttt{satisfied}.
\item Include \texttt{fix} only where \texttt{satisfied} is false. It must state the breaking case, the entailed constraint it violates, the code defect, and the concrete change --- describing behaviour and code without mentioning any requirement id.
\item Inside \texttt{fix} strings use plain prose; do not quote code with backticks or backslashes, which break JSON parsing.
\end{enumerate}

\textbf{Output:} Return exactly one JSON object:
\begin{flushleft}\small\texttt{%
\{"requirements": [\\
\hspace*{1.2em}\{"id": "\textless requirement id\textgreater", "satisfied": true\},\\
\hspace*{1.2em}\{"id": "\textless requirement id\textgreater", "satisfied": false, "fix": "..."\}\\
]\}}
\end{flushleft}
\end{PromptBox}

\begin{PromptBox}
{Refiner}
{Refiner prompt for critique-conditioned Web repair.}
{fig:prompt-refiner}

You are an expert frontend developer fixing a web page that failed automated testing.
A reviewer has flagged specific broken features, each with the defect and the fix.

Treat the existing code for every flagged feature as broken --- do not trust or copy it.
For each flagged defect, delete the current implementation of that feature and rewrite it from scratch so the described interaction works correctly: correct input validation, state updates, persistence, and DOM handling.
Keep the code only for features the reviewer did \emph{not} flag.

\textbf{Rules:}
\begin{enumerate}[nosep, leftmargin=*, itemsep=2pt]
\item For every flagged defect, rewrite that feature's JavaScript and DOM logic fresh; do not merely tweak the old code.
\item Preserve unflagged, working features as-is.
\item Output only the corrected HTML, from \texttt{\textless!DOCTYPE html\textgreater} to \texttt{\textless/html\textgreater}. No commentary, no markdown fences. Inline all CSS and JavaScript.
\end{enumerate}

\textbf{Input layout:} the prompt presents, in this order, the defect list to apply (\texttt{\{critique\}}), what the page must do (\texttt{\{input\_text\}}), and the current page to repair (\texttt{\{prev\_html\}}).
Foregrounding the defects before the previous page is deliberate: the refiner otherwise inherits a strong prior to copy its own earlier output and ignore the critique.

\textbf{Output:} the corrected, complete HTML page, with every flagged feature rewritten.
\end{PromptBox}

\end{document}